\makeatletter
\def\ReportTemplateLayout#1{%
  \begingroup\normalsize
  \typeout{TEMPLATE-LAYOUT #1 paperwidth=\the\paperwidth;paperheight=\the\paperheight;textwidth=\the\textwidth;textheight=\the\textheight;oddsidemargin=\the\oddsidemargin;evensidemargin=\the\evensidemargin;topmargin=\the\topmargin;parskip=\the\parskip;fontsize=\f@size;baselineskip=\the\baselineskip}%
  \ifx\@textbottom\relax
    \typeout{TEMPLATE-BOTTOM #1 flushbottom}%
  \else
    \typeout{TEMPLATE-BOTTOM #1 nondefault}%
  \fi
  \endgroup}
\makeatother
\AtBeginDocument{\ReportTemplateLayout{begin}}
\AtEndDocument{\ReportTemplateLayout{end}}

\documentclass{article} 
\ifdefined\pdfminorversion\pdfminorversion=7\fi
\usepackage{iclr2027_conference,times}

\usepackage{amsmath,amsfonts,bm}

\def\eqref#1{equation~\ref{#1}}

\def\1{\bm{1}}

\DeclareMathAlphabet{\mathsfit}{\encodingdefault}{\sfdefault}{m}{sl}
\SetMathAlphabet{\mathsfit}{bold}{\encodingdefault}{\sfdefault}{bx}{n}

\usepackage[utf8]{inputenc} 
\usepackage[T1]{fontenc}    
\usepackage{hyperref}       
\hypersetup{hidelinks}      
\usepackage{xurl}           
\usepackage{booktabs}       
\usepackage{longtable}      
\usepackage{tabularx}       
\usepackage[breakable,skins]{tcolorbox} 
\usepackage{needspace}      
\usepackage{amsfonts}       
\usepackage{nicefrac}       
\usepackage{microtype}      
\usepackage[table]{xcolor}  

\usepackage{subcaption}
\usepackage{multirow} 
\usepackage{makecell} 
\usepackage{float} 
\usepackage{placeins} 
\usepackage{xspace} 
\usepackage{enumitem} 
\usepackage{algorithm}
\usepackage{algorithmic}

\usepackage{amsmath}
\usepackage{amssymb}
\usepackage{mathtools}
\usepackage{amsthm}

\newcommand{\Tau}{\mathcal{T}}

\definecolor{impgreen}{RGB}{0,120,90}
\definecolor{impred}{RGB}{160,40,40}
\definecolor{agpBlue}{HTML}{3B69A1}
\definecolor{agpPurple}{HTML}{8060A4}

\usepackage[capitalize,noabbrev]{cleveref}
\theoremstyle{plain}
\newtheorem{theorem}{Theorem}[section]

\theoremstyle{definition}
\newtheorem{definition}[theorem]{Definition}

\theoremstyle{remark}

\newcommand{\projectname}{\textbf{\textsc{AGP}}\xspace}
\newcommand{\projectnameAgent}{\textbf{\textsc{AGS}}\xspace}

\title{Autogenesis: A Self-Evolving Agent Protocol}

\author{
    Wentao Zhang\textsuperscript{\rm 1}\footnotemark[1]\textbf{,}
    Zhe Zhao\textsuperscript{\rm 2}\footnotemark[1]\textbf{,}
    Haibin Wen\textsuperscript{\rm 3}\footnotemark[1]\textbf{,}
    Yingcheng Wu\textsuperscript{\rm 2}\textbf{,}
    Cankun Guo\textsuperscript{\rm 4}\textbf{,} \\
    \textbf{Ming Yin}\textsuperscript{\rm 2}\footnotemark[2]\textbf{,}
    \textbf{Bo An}\textsuperscript{\rm 1}\footnotemark[2] \\
    \textsuperscript{\rm 1} Nanyang Technological University \quad
    \textsuperscript{\rm 2} Stanford University \quad
    \textsuperscript{\rm 3} City University of Hong Kong \\
    \textsuperscript{\rm 4} University of Science and Technology of China \\
    \texttt{mingyin0312@gmail.com} \quad \texttt{boan@ntu.edu.sg}\\
    Project Code: \url{https://github.com/DVampire/Autogenesis}
}

\iclrfinalcopy 
\begin{document}

\maketitle
\lhead{Under review as a conference paper at ICLR 2027}
\begingroup
\renewcommand{\thefootnote}{\fnsymbol{footnote}}
\footnotetext[1]{Equal contribution.}
\footnotetext[2]{Corresponding authors.}
\endgroup

\begin{abstract}
LLM-based agent systems can adapt by revising prompts, constructing tools, and changing other runtime resources. Managing these heterogeneous updates requires a shared lifecycle for identifying targets, assessing candidates, and retaining accepted versions, complementing communication protocols such as A2A and MCP. We introduce \textbf{\textsc{Autogenesis Protocol (AGP)}}, which decouples what evolves from how updates are proposed. Its Resource Substrate Protocol Layer (RSPL) represents prompts, agents, tools, environments, and memory as protocol-registered resources\footnote{Resources instantiate the five RSPL types. Tools include local code-based tools, MCP tools, and skills. Outputs and solutions are task-local refinement targets, not persistent resource types.} with explicit state and versioned interfaces. Its Self-Evolution Protocol Layer (SEPL) specifies candidate proposal, assessment, and commit, with auditable histories and support for restoring retained resource states. We instantiate AGP in \textbf{\textsc{Autogenesis System (AGS)}}, a self-evolving multi-agent system that retrieves, instantiates, and refines resources during execution. Across reasoning, tool-use, and coding benchmarks, AGS improves over its non-evolving configuration, and repeated reasoning runs report the accompanying output-token cost. Execution cases show a shared update lifecycle across all five resource types.
\end{abstract}

\section{Introduction}

A long-running agent may repair a tool, revise an instruction, or change which memories survive compaction. Beyond generating a better answer, these adaptations require identifying the resource, evaluating the candidate, deciding when it becomes active, and retaining its history. A revised tool may change the evidence available to downstream reasoning even when its calling interface remains unchanged. We study a common contract for managing updates across resource types.

LLM-based systems combine reasoning, planning, and external tools~\citep{yao2022react,wei2022chain,wu2024autogen}. Communication standards address complementary concerns: MCP exposes tools, prompts, and contextual resources, while A2A supports agent discovery and task-oriented interaction~\citep{mcp2025spec,a2a2026spec}. We ask whether heterogeneous resource updates can share one lifecycle while each resource keeps its own execution contract. Our focus is a contract for proposing, evaluating, and retaining changes to heterogeneous agent-internal resources, complementing these communication mechanisms.

We propose \textbf{\textsc{Autogenesis Protocol (AGP)}}, with two complementary layers. The \textbf{Resource Substrate Protocol Layer (RSPL)} records resource identities, implementations, configurations, evolvability markers, and version histories. The \textbf{Self-Evolution Protocol Layer (SEPL)} composes proposal, evaluation, and commit operators over these records. An optimizer can thus propose changes through a shared lifecycle while resource-specific managers determine the permitted fields and invocation contracts. Together, the layers separate update proposals from the management of accepted resource versions and consumer bindings for subsequent task execution.

We instantiate AGP as \textbf{\textsc{Autogenesis System (AGS)}}, combining planner--specialist execution with reflection-driven resource updates. Our evaluation covers reasoning, tool use, and coding, complemented by repeated accuracy--cost comparisons and cases spanning all five resource types. These studies distinguish task-level refinement from protocol execution: benchmarks characterize system performance, while the cases illustrate concrete resource edits, acceptance decisions, and subsequent use by agents during task execution. We make three contributions:
\begin{itemize}[leftmargin=*]
    \item A two-layer resource-update contract separating heterogeneous resource management from the strategy used to propose changes.
    \item An AGS reference instantiation and three cases illustrating resource selection, candidate evaluation, versioned commit, and subsequent use across all five supported resource types.
    \item An empirical study on reasoning, tool-use, and coding benchmarks,\\
    including repeated runs with output-token accounting.
\end{itemize}

\section{Related Work}

\subsection{Agent Systems and Interoperability}
ReAct interleaves reasoning and actions, while AutoGen coordinates customizable agents through conversations~\citep{yao2022react,wu2024autogen}. MetaGPT encodes role-based operating procedures, and AgentVerse supports dynamically composed teams for collaborative problem solving~\citep{hong2023metagpt,chen2023agentverse}. MCP standardizes contextual resources, prompts, tools, and client--server interaction, while A2A standardizes discovery, messages, tasks, and artifacts~\citep{mcp2025spec,a2a2026spec}. AGP complements these mechanisms by managing candidate changes, activation decisions, and retained versions of the underlying resources. For example, a tool exposed through MCP can retain its communication interface while AGP manages updates to its implementation. The communication contract specifies how the tool is invoked; the update contract specifies which implementation is available to subsequent consumers. Appendix~\ref{appx_sec:comparison-with-other-protocols} compares these objects and lifecycle boundaries.

\subsection{Refinement and Resource Optimization}
Self-Refine uses an LLM's feedback to revise its current response, whereas Reflexion stores verbal reflections to guide subsequent trials~\citep{madaan2023self,shinn2023reflexion}. TextGrad propagates textual feedback through computation graphs; MIPRO optimizes module instructions and demonstrations; GEPA uses trajectory reflection to guide prompt search~\citep{yuksekgonul2025optimizing,opsahlong2024mipro,agrawal2026gepa}. AFlow and EvoAgentX extend optimization to workflow structure~\citep{zhang2025aflow,wang2025evoagentx}, while Alita constructs tools that can be reused in later execution~\citep{qiu2025alita}. These methods already retain feedback, candidates, or reusable artifacts. AGP specifies a shared contract for resource identity, candidate evaluation, and versioned commit across resource types. Appendix~\ref{app:optimization-comparison} compares update targets, feedback, scope, and retained state. This comparison concerns mechanisms, not matched-budget performance.

\section{Autogenesis Protocol}

\begin{figure*}[t]

  \centering
  \includegraphics[width=0.98\textwidth]{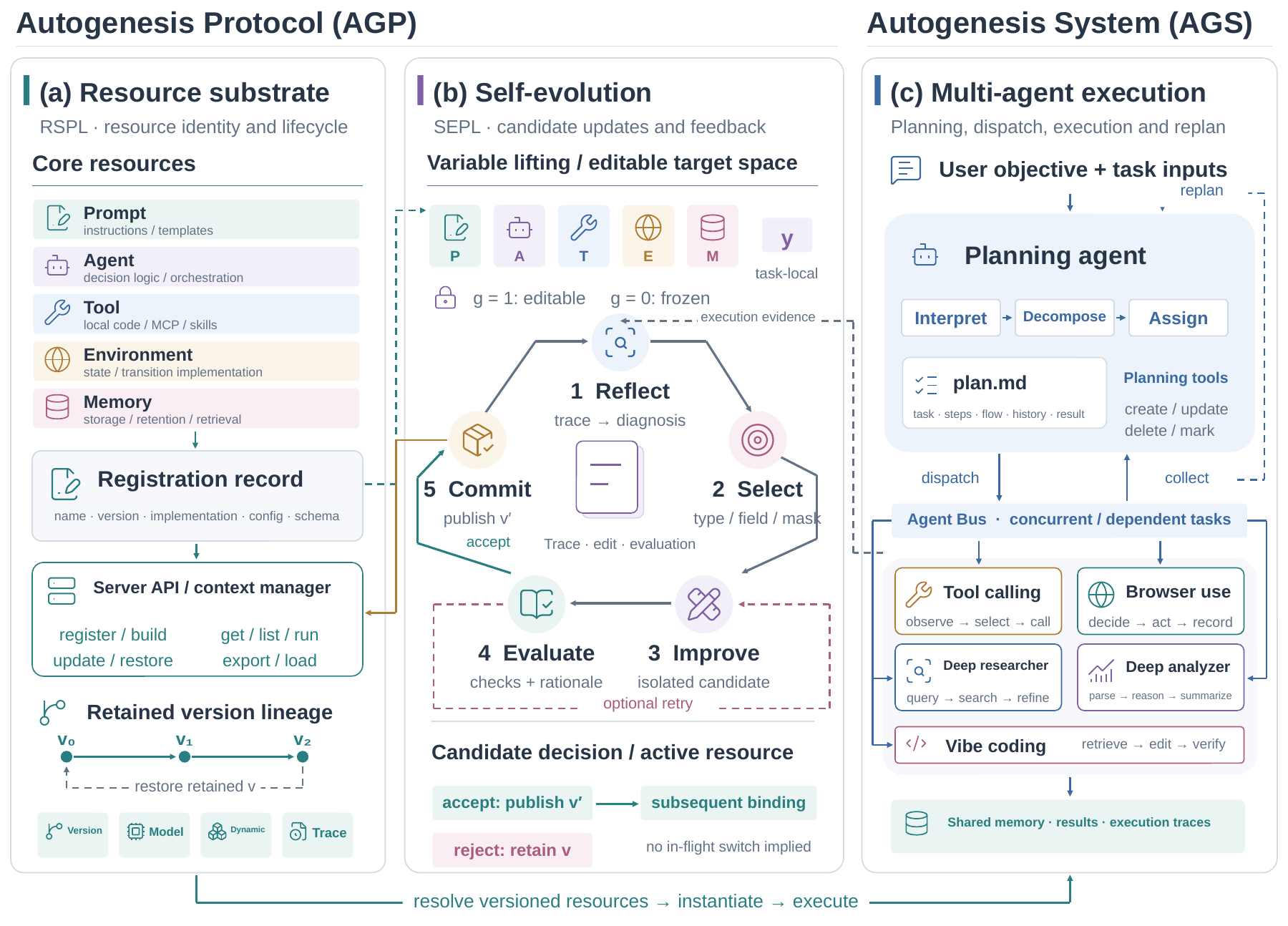}
  \caption{Autogenesis architecture. (a) RSPL manages five resource types and retained versions. (b) SEPL organizes candidate updates, evaluation, and commit; rejection differs from restoring a retained version. (c) AGS coordinates planning, specialists, and shared memory. Execution traces inform updates, and subsequent execution binds versioned resources. The version path is schematic.}
  \label{fig:architecture}
\end{figure*}

Figure~\ref{fig:architecture} links task execution, candidate updates, and resource management. In the spreadsheet case, the Tool \texttt{xlsx\_reader\_tool} is registered at version \texttt{v1.0.0}: its implementation reads a workbook, and its invocation contract returns rows, path, and sheet. File-derived checks precede commit; the aggregation Agent then invokes the accepted Tool through its registered interface. Later Agent and Prompt changes follow the same lifecycle (Appendix~\ref{app:case-tool-agent-prompt}). RSPL records identities and versions; SEPL coordinates proposal, assessment, and commit across resource types.

\subsection{Layer 1: Resource Substrate Protocol Layer}

RSPL supports five resource types: instructions (\textsc{Prompt}), decision logic (\textsc{Agent}), callable capabilities (\textsc{Tool}), executable environment models (\textsc{Environment}), and storage and retrieval policies (\textsc{Memory}). Tools include local implementations, MCP-connected capabilities, and skills. SEPL acts on the exposed implementation or configuration fields through management interfaces; ordinary task-local state changes remain part of execution. This separation allows the same representation to support different update strategies without prescribing an exhaustive resource taxonomy.

\begin{samepage}
\begin{definition}[Resource Entity]
\label{def:resource-entity}
A resource entity and its type-level collection are defined as:
\begin{equation}
\begin{aligned}
e_{\tau,i} &= (n_{\tau,i},\, d_{\tau,i},\, \phi_{\tau,i},\, g_{\tau,i},\, m_{\tau,i}),\\
\mathcal{E}_{\tau} &= \{\, e_{\tau,i} \mid i \in \mathcal{I}_{\tau} \,\},
\end{aligned}
\end{equation}
where $\Tau = \{\textsc{Prompt},\textsc{Agent},\textsc{Tool},\textsc{Env},\textsc{Mem}\}$ indexes types and $i\in\mathcal{I}_\tau$ indexes registered resources of type $\tau$. The name $n_{\tau,i}$ is unique within its type, giving the typed identity $(\tau,n_{\tau,i})$. Other fields are a description $d_{\tau,i}$, invocation mapping $\phi_{\tau,i}:\mathcal{X}_{\tau}\rightarrow\mathcal{Y}_{\tau}$, evolvability marker $g_{\tau,i}\in\{0,1\}$, and metadata dictionary $m_{\tau,i}$. Invocation mappings include relevant runtime state.
\end{definition}
\end{samepage}

\begin{samepage}
\begin{definition}[Resource Registration Record]
  \label{def:resource-registration-record}
  The record and its type-level collection are
  \begin{equation}
  \begin{aligned}
  c_{\tau,i} &= (e_{\tau,i},\, v_{\tau,i},\, \eta_{\tau,i},\, \theta_{\tau,i},\, \mathcal{F}_{\tau,i}),\\
  \mathcal{C}_{\tau} &= \{\, c_{\tau,i} \mid i \in \mathcal{I}_{\tau} \,\},
  \end{aligned}
  \end{equation}
  where $e_{\tau,i}$ is the entity from \cref{def:resource-entity}, $v_{\tau,i}\in\mathbb{V}$ identifies its version, $\eta_{\tau,i}$ specifies its implementation, and $\theta_{\tau,i}$ supplies instantiation parameters. The exported representations $\mathcal{F}_{\tau,i}$ describe how consumers interact with the resource, including function-calling schemas and text.
  \end{definition}
\end{samepage}

\begin{definition}[Type-level resource interface]
\label{def:protocol-registered-resource}
For type $\tau$, the registry $\mathcal{R}_\tau$ maps typed identities to active records in $\mathcal{C}_\tau$; retained versions are managed separately. The global registry $\mathcal{R}=\bigcup_\tau\mathcal{R}_\tau$ uses these type-qualified keys. RSPL exposes a type-level management interface, rather than a single resource instance, through the triple
\begin{equation}
r_{\tau} = (\mathcal{C}_\tau,\; \mathcal{M}_\tau,\; \mathcal{A}_\tau),
\end{equation}
where  $\mathcal{C}_\tau$ stores registration records, $\mathcal{M}_\tau$ manages registration, construction, execution, update, and restoration, and $\mathcal{A}_\tau$ exposes these operations to consumers. Exported contracts specify capabilities and arguments. Retained histories cover resource configurations and associated artifacts rather than arbitrary external state; Appendix~\ref{appx_sec:context-manager} details the APIs and their restoration boundaries.
\end{definition}

\subsubsection{Infrastructure Services}

Four infrastructure services support resource management. The \textbf{model manager} routes requests across configured providers. The \textbf{version manager} tracks identifiers and lineage; resource managers retain configurations and artifacts used for restoration. The \textbf{dynamic manager} loads implementations and converts schemas, while the \textbf{trace manager} links execution observations to proposed changes and evaluation outcomes. Together, these services connect an update's motivating evidence to its recorded version and subsequent use, with activation and restoration boundaries specified below.

\subsection{Layer 2: Self-Evolution Protocol Layer (SEPL)}

SEPL organizes updates as a closed loop over candidate states and evaluation messages. Here, self-evolution refers to feedback-driven edits to resources and task-local solutions rather than model-parameter training. In the reflection instantiation used by our experiments, execution evidence motivates an edit, evaluation assesses the candidate, and commit records an accepted version. The interface separates this proposal strategy from RSPL's management operations. Appendix~\ref{sec:appendix-self-evolution-protocol-layer} provides the reflection algorithm and extension sketches.

\subsubsection{Evolvable Variables}

\emph{Variable lifting} resolves a resource entity to editable fields of its registration record. The marker restricts targets; the manager specifies permitted fields, such as implementation $\eta$ or parameters $\theta$. Task-local outputs are also eligible for discrete, non-differentiable candidate edits.

\begin{definition}[Evolvable Variable Set]
  We write $\mathcal{V}_{\text{evo}}=(\bigcup_{\tau\in\Tau}\mathcal{E}_{\tau})\cup\{y\}$, where $y$ is a task-local output/solution, not a sixth persistent resource type. A marker $g_v\in\{0,1\}$ selects $\Theta=\{v\in\mathcal{V}_{\text{evo}}\mid g_v=1\}$. The state space $\mathfrak{V}$ includes configurations with new resource instances. Traces remain observational evidence and cannot be rewritten as candidate resource edits.
\end{definition}

\subsubsection{Operator Composition}

SEPL operators carry resource/output states alongside auxiliary messages containing candidates. The protocol requires proposal and assessment to leave active resource configurations and bindings unchanged; only accepted commits may publish updates. Composition requires compatible message types and lifecycle preconditions.

\begin{definition}[SEPL Operator]
\label{def:sepl-operator}
Let $\mathfrak{V}$ be the state space and $\mathcal{P}$ the message space carrying traces, hypotheses, edits, candidates, and assessments. A \emph{SEPL operator} has the interface
\begin{equation}
f: \mathfrak{V} \times \mathcal{P}_{\text{in}} \;\rightarrow\; \mathfrak{V} \times \mathcal{P}_{\text{out}},
\end{equation}
where $\mathcal{P}_{\text{in}},\mathcal{P}_{\text{out}}\subseteq\mathcal{P}$ specify message types. Applied to a state and message, $f$ returns $(\mathcal{V}'_{\text{evo}},p')$ with $\mathcal{V}'_{\text{evo}}\in\mathfrak{V}$ and $p'\in\mathcal{P}_{\text{out}}$. Compatible operators compose as $f_n\circ\cdots\circ f_1$. This is a dataflow interface; sampled proposals may be stochastic, while recovery requires consistent retained state.
\end{definition}

\subsubsection{Evolutionary Loop}

Initial execution under $\mathcal{V}_{\text{evo}}^{(0)}\in\mathfrak{V}$ produces the trace carried by $p^{(0)}\in\mathcal{P}$. At iteration $t$, compatible operators transform the current state and its associated message as follows:
\begin{equation}
\bigl(\mathcal{V}_{\text{evo}}^{(t+1)},\, p^{(t+1)}\bigr) = (f_n \circ \cdots \circ f_1)\bigl(\mathcal{V}_{\text{evo}}^{(t)},\, p^{(t)}\bigr),
\end{equation}
where the reflection cycle comprises \textsc{Reflect}, \textsc{Select}, \textsc{Improve}, \textsc{Evaluate}, and \textsc{Commit}, with task continuation only when required. Rejection must preserve the active resource binding while its assessment informs later proposals. Accepted solution edits retain task-local $y$ without creating a persistent resource or triggering an unconditional new answer. Appendix~\ref{sec:appendix-self-evolution-protocol-layer} specifies these rules.

\paragraph{Candidate, commit, and restoration boundaries.}
Commit records an accepted version for subsequent use; publication and consumer binding are distinct events. Restoration must realign retained configurations, version records, and consumer bindings. It assumes mediated mutations and a trusted registry/version store, covers only captured internal state, and cannot undo completed external actions. The cases in Appendix~\ref{app:cases} exercise accepted updates, not recovery after faulty activation.

\paragraph{Feedback boundaries.}
For GPQA-Diamond/AIME, evaluation is a separate call to the same backbone, receiving the question and candidate reasoning/answer and returning a satisfaction flag and rationale. Reference answers are used only for final scoring. Executable tasks additionally provide execution feedback. Appendix~\ref{app:feedback-boundary} lists the optimizer-visible fields; a positive self-assessment expresses the evaluator's judgment, not correctness against the withheld reference answer.

\section{Autogenesis System}
\label{sec:system}

AGS instantiates AGP with a planner and five specialist roles. This organization builds on prior multi-agent and planner--specialist frameworks~\citep{wu2024autogen,skywork2025superagents}; an Agent Bus handles dispatch and result collection. The bus is an implementation choice, not a protocol requirement. Resource identities remain separate from instantiated implementations, allowing subsequent dispatches to resolve an accepted version. The resulting workflow proceeds as follows.

\textbf{Task planning.} The planner decomposes each incoming task without executing the subtasks itself. It maintains a structured \texttt{plan.md} containing the original task, assigned subtask steps, an execution flowchart, execution history, and a final result summary. Subtasks are assigned to deep researcher, browser-use, deep analyzer, tool-calling, or coding agents. The planner dispatches independent subtasks concurrently and respects dependencies between sequential steps, collecting specialist outputs through the bus before revising the plan and determining dependencies for the next round.

\textbf{Execution and replanning.} Each specialist retrieves the relevant prompt and tool resources from RSPL, executes its subtask, and writes results and traces to shared memory. After collecting the outputs, the planner updates \texttt{plan.md} and either concludes the task or initiates another round of decomposition. AGS also supports \emph{agent-as-tool} composition: a specialist is exposed through an RSPL tool schema and invoked directly by another agent. This alternative uses the same resource interfaces without requiring bus-level orchestration, using the registered contracts for tool invocation.

\textbf{Resource updates.} Execution traces can trigger the reflection loop in Algorithm~\ref{alg:sepl-loop-appx}. The optimizer selects editable resources, constructs candidates, and evaluates them before commit. Subsequent execution can then use an accepted version through RSPL. The cases distinguish this post-commit execution from candidate replay; they also record when a frozen implementation requires creating a new resource rather than modifying the existing one. Publication does not automatically switch every in-flight consumer, so the binding scope remains part of the execution record.

The evaluated AGS update strategies use reflection. Appendix~\ref{sec:appendix-self-evolution-protocol-layer} presents unevaluated textual-feedback and reward-informed sketches as partial mappings. Full integration requires candidate assessment and acceptance-gated publication. The reward-informed sketches use text similarity, not token likelihood ratios, and are not standard Reinforce++ or GRPO training.

\section{Empirical Studies}

\begin{figure}[t]
\centering
\includegraphics[width=\textwidth]{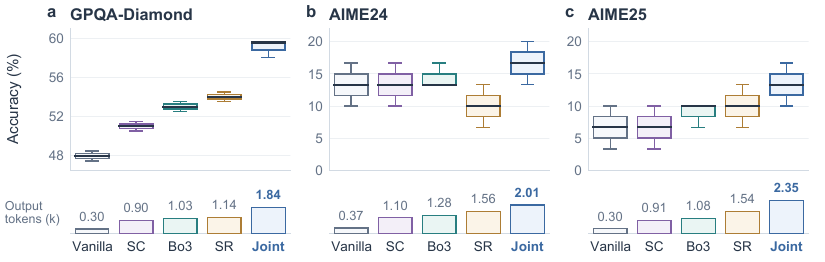}
\caption{GPT-4o accuracy and cost (three runs). Boxes show the IQR and median; whiskers show the range. Bars show output tokens per problem (k). Appendix~\ref{app:repeated-controls} gives exact results and settings.}
\label{fig:opportunity-controls}
\end{figure}

We ask three questions: (i) How do different edit targets affect reasoning accuracy, and what output-token costs accompany the repeated evaluations (Table~\ref{tab:results1}; Figure~\ref{fig:opportunity-controls})? (ii) What task-level performance do the configured pipelines achieve in tool use and coding (Table~\ref{tab:results2}, Figure~\ref{fig:hle}, and Table~\ref{tab:results3})? (iii) How do different resource types follow the same update lifecycle (Appendix~\ref{app:cases})? The benchmarks characterize task outcomes, while the cases illustrate resource-management behavior.

\textbf{Benchmarks.} GPQA-Diamond has 198 graduate-level multiple-choice questions; AIME24 and AIME25 each have 30 integer-answer problems. Reasoning uses no external retrieval. GAIA~\citep{mialon2023gaia} tests planning and tool use across three difficulty levels; Humanity's Last Exam (HLE)~\citep{phan2025humanity} covers expert questions. Coding evaluation uses 100 selected LeetCode problems in Python3, C++, Java, Go, and Kotlin, reporting acceptance, execution statistics, and submission-relative metrics. Appendix~\ref{app:coding-benchmark} details the setup, metrics, and reproducibility constraints.

\begin{table*}[t]
\footnotesize
\begin{minipage}[t]{0.479\textwidth}
\centering
\footnotesize
\renewcommand{\arraystretch}{1.26745}
\setlength{\tabcolsep}{0pt}
\caption{Reasoning accuracy (\%) with counts.}
\label{tab:results1}
\begin{tabular*}{\linewidth}{@{\extracolsep{\fill}}lrrr@{}}
\toprule
\textbf{Agent} & {\scriptsize\textbf{GPQA-Diamond}} & {\scriptsize\textbf{AIME24}} & {\scriptsize\textbf{AIME25}} \\
\midrule
\rowcolor{agpBlue!7} \multicolumn{4}{c}{\textit{\textbf{gpt-4o}}} \\
Vanilla & 47.98\,{\scriptsize(95/198)} & 13.33\,{\scriptsize(4/30)} & 6.67\,{\scriptsize(2/30)} \\
Prompt-Evo & 53.54\,{\scriptsize(106/198)} & 13.33\,{\scriptsize(4/30)} & 13.33\,{\scriptsize(4/30)} \\
Solution-Evo & 53.54\,{\scriptsize(106/198)} & 16.67\,{\scriptsize(5/30)} & 13.33\,{\scriptsize(4/30)} \\
PS-Joint-Evo & 58.08\,{\scriptsize(115/198)} & 16.67\,{\scriptsize(5/30)} & 13.33\,{\scriptsize(4/30)} \\
\cmidrule{2-4}
\textbf{Gain (\%)} & \textbf{21.05$\uparrow$} & \textbf{25.00$\uparrow$} & \textbf{100.00$\uparrow$} \\
\midrule
\rowcolor{agpBlue!7} \multicolumn{4}{c}{\textit{\textbf{gpt-4.1}}} \\
Vanilla & 65.15\,{\scriptsize(129/198)} & 23.33\,{\scriptsize(7/30)} & 20.00\,{\scriptsize(6/30)} \\
Prompt-Evo & 68.69\,{\scriptsize(136/198)} & 33.33\,{\scriptsize(10/30)} & 23.33\,{\scriptsize(7/30)} \\
Solution-Evo & 68.69\,{\scriptsize(136/198)} & 36.67\,{\scriptsize(11/30)} & 30.00\,{\scriptsize(9/30)} \\
PS-Joint-Evo & 67.68\,{\scriptsize(134/198)} & 40.00\,{\scriptsize(12/30)} & 33.33\,{\scriptsize(10/30)} \\
\cmidrule{2-4}
\textbf{Gain (\%)} & \textbf{3.88$\uparrow$} & \textbf{71.43$\uparrow$} & \textbf{66.67$\uparrow$} \\
\midrule
\rowcolor{agpBlue!7} \multicolumn{4}{c}{\textit{\textbf{grok-4.1-fast}}} \\
Vanilla & 83.33\,{\scriptsize(165/198)} & 96.67\,{\scriptsize(29/30)} & 90.00\,{\scriptsize(27/30)} \\
Prompt-Evo & 83.84\,{\scriptsize(166/198)} & 96.67\,{\scriptsize(29/30)} & 93.33\,{\scriptsize(28/30)} \\
Solution-Evo & 87.88\,{\scriptsize(174/198)} & 96.67\,{\scriptsize(29/30)} & 90.00\,{\scriptsize(27/30)} \\
PS-Joint-Evo & 89.39\,{\scriptsize(177/198)} & 96.67\,{\scriptsize(29/30)} & 96.67\,{\scriptsize(29/30)} \\
\cmidrule{2-4}
\textbf{Gain (\%)} & \textbf{7.27$\uparrow$} & 0.00 & \textbf{7.41$\uparrow$} \\
\midrule
\rowcolor{agpBlue!7} \multicolumn{4}{c}{\textit{\textbf{claude-sonnet-4.5}}} \\
Vanilla & 78.28\,{\scriptsize(155/198)} & 76.67\,{\scriptsize(23/30)} & 73.33\,{\scriptsize(22/30)} \\
Prompt-Evo & 79.80\,{\scriptsize(158/198)} & 86.67\,{\scriptsize(26/30)} & 90.00\,{\scriptsize(27/30)} \\
Solution-Evo & 80.30\,{\scriptsize(159/198)} & 80.00\,{\scriptsize(24/30)} & 90.00\,{\scriptsize(27/30)} \\
PS-Joint-Evo & 81.31\,{\scriptsize(161/198)} & 86.67\,{\scriptsize(26/30)} & 90.00\,{\scriptsize(27/30)} \\
\cmidrule{2-4}
\textbf{Gain (\%)} & \textbf{3.87$\uparrow$} & \textbf{13.04$\uparrow$} & \textbf{22.73$\uparrow$} \\
\midrule
\rowcolor{agpBlue!7} \multicolumn{4}{c}{\textit{\textbf{gemini-3-flash-preview}}} \\
Vanilla & 88.38\,{\scriptsize(175/198)} & 83.33\,{\scriptsize(25/30)} & 83.33\,{\scriptsize(25/30)} \\
Prompt-Evo & 88.89\,{\scriptsize(176/198)} & 93.33\,{\scriptsize(28/30)} & 86.67\,{\scriptsize(26/30)} \\
Solution-Evo & 87.88\,{\scriptsize(174/198)} & 93.33\,{\scriptsize(28/30)} & 90.00\,{\scriptsize(27/30)} \\
PS-Joint-Evo & 90.40\,{\scriptsize(179/198)} & 93.33\,{\scriptsize(28/30)} & 93.33\,{\scriptsize(28/30)} \\
\cmidrule{2-4}
\textbf{Gain (\%)} & \textbf{2.29$\uparrow$} & \textbf{12.00$\uparrow$} & \textbf{12.00$\uparrow$} \\
\bottomrule
\end{tabular*}

\end{minipage}%
\hfill%
\begin{minipage}[t]{0.505\textwidth}
\centering
\footnotesize
\newcommand{\gaiacell}[2]{\mbox{#1 {\scriptsize#2}}}
\renewcommand{\arraystretch}{1.33294}
\setlength{\tabcolsep}{0.4pt}
\caption{GAIA scores (\%).}
\label{tab:results2}
\begin{tabular*}{\linewidth}{@{\extracolsep{\fill}}lrrrr@{}}
\toprule
\textbf{Agent} & \textbf{L1} & \textbf{L2} & \textbf{L3} & \textbf{Avg.} \\
\midrule
\rowcolor{agpBlue!7} \multicolumn{5}{c}{\textit{\textbf{Validation}}} \\
\gaiacell{HF ODR}{\citep{huggingface_open_deep_research_2024}} & 67.92 & 53.49 & 34.62 & 55.15 \\
\gaiacell{o3-DR}{\citep{openai2025deepresearch}} & 74.29 & 69.06 & 47.60 & 67.36 \\
\gaiacell{DeSearch}{\citep{desearchai2024desearch}} & 90.57 & 72.01 & 38.46 & 72.73 \\
\gaiacell{Co-Sight}{\citep{zhang2025cosight}} & 86.79 & 73.26 & 42.31 & 72.73 \\
\gaiacell{Manus}{\citep{shen2025mindmachinerisemanus}} & 86.50 & 70.10 & 57.69 & 73.90 \\
\gaiacell{AWorld}{\citep{yu2025aworld}} & 88.68 & 77.91 & 53.85 & 77.58 \\
\gaiacell{Langfun}{\citep{google2024langfun}} & 88.68 & 80.23 & 57.69 & 79.39 \\
\gaiacell{Skywork}{\citep{skywork2025superagents}} & 92.45 & 83.72 & 57.69 & 82.42 \\
\gaiacell{agent-2030}{\citep{gaia2025results}} & 96.23 & 90.70 & 57.69 & 87.27 \\
\gaiacell{Alita\smash{$^{\dagger}$}}{\citep{qiu2025alita}} & 88.68 & 89.53 & 76.92 & 87.27 \\
\midrule
Vanilla                                  & 92.45 & 88.37 & 88.46 & 89.70 \\
Agent-Evo                             & \textbf{96.23} & \textbf{93.02} & \textbf{88.46} & \textbf{93.33} \\
\cmidrule{2-5}
\textbf{Gain (\%)} & \textbf{4.09$\uparrow$} & \textbf{5.26$\uparrow$} & 0.00 & \textbf{4.05$\uparrow$} \\
\midrule

\rowcolor{agpBlue!7} \multicolumn{5}{c}{\textit{\textbf{Test}}} \\
\gaiacell{o4-mini-DR}{\citep{openai2025deepresearch}} & 67.59 & 59.10 & 44.28 & 59.30 \\
\gaiacell{JoyAgent}{\citep{liu2025joyagent}} & 77.42 & 67.30 & 46.94 & 67.11 \\
\gaiacell{o3-DR}{\citep{openai2025deepresearch}} & 79.42 & 68.97 & 47.48 & 68.70 \\
\gaiacell{Langfun}{\citep{google2024langfun}} & 84.95 & 73.58 & 48.98 & 73.09 \\
\gaiacell{Alita}{\citep{qiu2025alita}} & 92.47 & 71.70 & 55.10 & 75.42 \\
\gaiacell{DeSearch}{\citep{desearchai2024desearch}} & 91.40 & 75.47 & 61.22 & 78.07 \\
\gaiacell{h2o}{\citep{h2oai2025h2ogpte}} & 89.25 & 79.87 & 61.22 & 79.73 \\
\gaiacell{Su-Zero-Ultra}{\citep{gaia2025results}} & 93.55 & 77.36 & 65.31 & 80.40 \\
\gaiacell{AWorld}{\citep{yu2025aworld}} & 95.70 & 81.13 & 57.14 & 81.73 \\
\gaiacell{HALO}{\citep{hou2025halo}} & 94.62 & 84.91 & 69.39 & 85.38 \\
\gaiacell{ToolOrchestra}{\citep{su2025toolorchestra}} & 95.70 & 82.39 & 87.76 & 87.38 \\
\gaiacell{openJiuwen}{\citep{openjiuwen2025}} & \textbf{98.92} & \textbf{88.68} & \textbf{87.76} & \textbf{91.69} \\
\midrule
Vanilla                                  & 91.40 & 77.36 & 61.22 & 79.07 \\
Agent-Evo                             & 98.92 & 85.53 & 81.63 & 89.04 \\
\cmidrule{2-5}
\textbf{Gain (\%)} & \textbf{8.23$\uparrow$} & \textbf{10.56$\uparrow$} & \textbf{33.34$\uparrow$} & \textbf{12.61$\uparrow$} \\
\bottomrule
\end{tabular*}
\end{minipage}
\par\smallskip
{\footnotesize\raggedright Table~\ref{tab:results1}: single complete evaluations; accuracy and Gain use integer counts before rounding. Repeated controls are separate (Appendix~\ref{app:repeated-controls}). Table~\ref{tab:results2}: external systems use their own backbones and budgets; $\dagger$Alita validation uses pass@3. Gain compares the AGS variants only (Appendix~\ref{app:gaia-sources}).\par}
\par\smallskip\noindent
\begin{minipage}[t]{\textwidth}
\centering
\includegraphics[width=\textwidth, height=0.1\textheight]{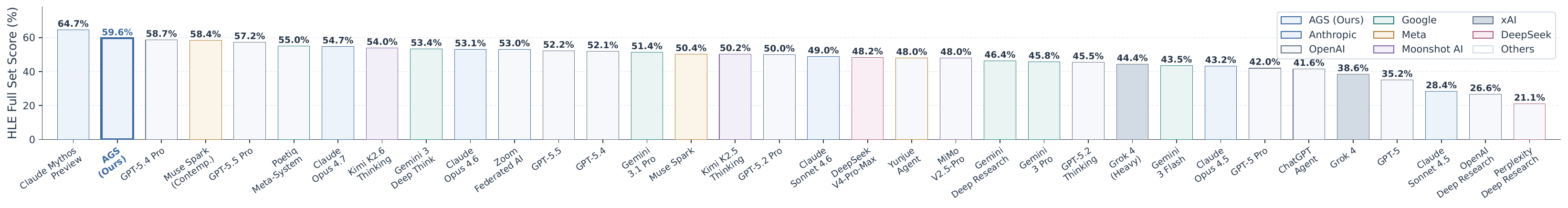}
\captionof{figure}{HLE full-set scores of agent systems~\citep{zoom2025hleleaderboard}.}
\label{fig:hle}
\end{minipage}
\end{table*}

\subsection{Experiments on Scientific and Mathematical Benchmarks}

\textbf{Experimental setup.} We evaluate \projectnameAgent on GPQA-Diamond, AIME24, and AIME25. \emph{Prompt-Evo} edits prompts, \emph{Solution-Evo} performs within-task solution refinement, and \emph{PS-Joint-Evo} jointly edits both. The multi-model comparison permits up to three reflection rounds without external tools and scores the final output after termination. Accuracy requires the correct GPQA-Diamond option or exact AIME integer. Table entries are single runs; repeated controls are analyzed separately. Reference answers are used only for scoring (Appendix~\ref{app:feedback-boundary}).

\textbf{Results.} Joint evolution improves on Vanilla in most reported configurations, although the relative benefit of editing both targets depends on the backbone and task. On GPT-4.1 GPQA-Diamond, joint evolution scores 67.68 compared with 68.69 for each single-target variant; on Grok-4.1-fast AIME24, all evolution variants score 96.67. Net GPQA-Diamond gains over Vanilla span 4--20 correct answers across backbones. On AIME, each additional correct answer changes accuracy by 3.33 percentage points, making counts essential to interpreting relative gains. Joint editing is therefore not uniformly preferable, and its benefit should be weighed against inference cost. We supplement these single-run comparisons with repeated GPT-4o evaluations that report per-run counts, accuracy variation, and output-token costs under controlled primary generation opportunities.

\paragraph{Repeated, opportunity-controlled comparison.}
We additionally compare GPT-4o under at most three primary generation/refinement opportunities. Vanilla uses one unrefined answer; Self-Consistency (SC) samples three answers and votes; Best-of-3 (Bo3) uses a model judge; Simple Reflection (SR) produces an initial solution and two refinements without persistent resource updates; PS-Joint-Evo (Joint) edits the prompt and solution. All use the same reported route and decoding configuration, with a 32,768 output-token ceiling per call. This ceiling is not an equal realized-compute budget. Auxiliary reflection, editing, and evaluation calls must be distinguished from primary opportunities when interpreting the measured accuracy--cost trade-off.

Across three evaluations, joint editing reaches 59.09\% mean accuracy on GPQA-Diamond versus 54.04\% for Simple Reflection (351/594 versus 321/594 correct responses). On AIME24, it yields 15/90 versus 13/90 for Best-of-3; on AIME25, 12/90 versus 9/90 for Simple Reflection. Gains are 5.05, 2.22, and 3.33 percentage points, with approximately 61\%, 57\%, and 52\% more output tokens than these respective controls. Denominators count repeated responses to the same 198 or 30 questions, not additional independent problems. These results characterize a higher-accuracy operating point at increased output-token cost, not an equal-compute advantage.

\subsection{Experiments on General Agent Benchmarks}

\begin{table}[!t]
\centering
\scriptsize
\caption{Coding results on 100 LeetCode problems. Gain compares Solution-Evo with the preceding Gemini baseline ($\uparrow$: favorable; $\downarrow$: unfavorable). Appendix~\ref{app:coding-benchmark} defines metrics and aggregation.}
\label{tab:results3}
\setlength{\tabcolsep}{1.3pt}
\renewcommand{\arraystretch}{0.96}
\setlength{\aboverulesep}{0.25ex}
\setlength{\belowrulesep}{0.35ex}
\begin{tabular}{@{}l*{13}{r}@{}}
\toprule
\textbf{Model} & \multicolumn{8}{c}{\textbf{Capability metrics}} & \multicolumn{3}{c}{\textbf{Execution metrics}} & \multicolumn{2}{c}{\textbf{Relative metrics}} \\
\cmidrule(lr){2-9}\cmidrule(lr){10-12}\cmidrule(lr){13-14}
 & \textbf{PR} & \textbf{TLE} & \textbf{MLE} & \textbf{CE} & \textbf{RE} & \textbf{WA} & \textbf{TO} & \textbf{RpE} & \makecell{\textbf{AR (ms)}} & \makecell{\textbf{AM (MB)}} & \makecell{\textbf{APC}} & \makecell{\textbf{ARB (\%)}} & \multicolumn{1}{c}{\textbf{AMB (\%)}} \\

\midrule
\rowcolor{agpBlue!7} \multicolumn{14}{c}{\textit{\textbf{Python3}}} \\
deepseek-v3.2 & 34 & 1 & 1 & 0 & 23 & 8 & 1 & 36 & 1806.79 & 55.91 & 640.38 & 63.04 & 25.81 \\
grok-4.1-fast & 73 & 9 & 0 & 0 & 3 & 13 & 0 & 3 & 1860.90 & 56.02 & 741.60 & 49.92 & 30.15 \\
claude-sonnet-4.5 & 42 & 37 & 2 & 0 & 0 & 10 & 8 & 1 & 880.98 & 45.16 & 702.64 & 61.06 & 22.12 \\
claude-opus-4.5 & 82 & 9 & 0 & 0 & 0 & 5 & 3 & 1 & 1559.87 & 70.77 & 749.45 & 64.77 & 32.70 \\
gemini-3-flash-preview & 79 & 4 & 0 & 0 & 2 & 14 & 1 & 0 & 1376.19 & 56.59 & 750.89 & 73.28 & 36.62 \\
\rowcolor{agpPurple!8}\quad + Solution-Evo & 87 & 3 & 0 & 0 & 1 & 9 & 0 & 0 & 1269.39 & 59.08 & 750.98 & 70.29 & 42.15 \\
\cmidrule{2-14}
\textbf{Gain (\%)} & \textbf{10.1$\uparrow$} & \textbf{25.0$\uparrow$} & 0 & 0 & \textbf{50$\uparrow$} & \textbf{35.7$\uparrow$} & \textbf{100$\uparrow$} & 0 & \textbf{7.8$\uparrow$} & 4.4$\downarrow$ & 0.0 & 4.1$\downarrow$ & 15.1$\uparrow$ \\

\midrule
\rowcolor{agpBlue!7} \multicolumn{14}{c}{\textit{\textbf{C++}}} \\
deepseek-v3.2 & 11 & 1 & 0 & 30 & 0 & 6 & 4 & 43 & 158.73 & 163.59 & 605.82 & 73.11 & 74.05 \\
grok-4.1-fast & 79 & 9 & 0 & 1 & 0 & 5 & 2 & 4 & 428.32 & 223.68 & 748.61 & 58.57 & 46.67 \\
claude-sonnet-4.5 & 41 & 42 & 2 & 0 & 1 & 9 & 2 & 3 & 379.68 & 179.86 & 710.59 & 56.17 & 50.84 \\
claude-opus-4.5 & 85 & 6 & 0 & 0 & 0 & 6 & 1 & 2 & 382.45 & 184.22 & 758.21 & 64.06 & 55.58 \\
gemini-3-flash-preview & 84 & 2 & 0 & 2 & 1 & 10 & 0 & 1 & 266.04 & 168.93 & 743.31 & 68.02 & 59.24 \\
\rowcolor{agpPurple!8}\quad + Solution-Evo & 99 & 0 & 0 & 0 & 0 & 1 & 0 & 0 & 142.60 & 148.43 & 749.86 & 88.99 & 73.14 \\
\cmidrule{2-14}
\textbf{Gain (\%)} & \textbf{17.9$\uparrow$} & \textbf{100$\uparrow$} & 0 & \textbf{100$\uparrow$} & \textbf{100$\uparrow$} & \textbf{90$\uparrow$} & 0 & \textbf{100$\uparrow$} & \textbf{46.4$\uparrow$} & \textbf{12.1$\uparrow$} & \textbf{0.9$\uparrow$} & \textbf{30.8$\uparrow$} & \textbf{23.5$\uparrow$} \\

\midrule
\rowcolor{agpBlue!7} \multicolumn{14}{c}{\textit{\textbf{Java}}} \\
deepseek-v3.2 & 11 & 0 & 0 & 47 & 1 & 1 & 5 & 35 & 72.91 & 143.63 & 481.45 & 57.37 & 32.71 \\
grok-4.1-fast & 73 & 5 & 0 & 5 & 0 & 12 & 1 & 4 & 227.45 & 136.80 & 746.23 & 52.98 & 41.97 \\
claude-sonnet-4.5 & 41 & 40 & 1 & 1 & 1 & 15 & 0 & 1 & 161.49 & 130.54 & 679.41 & 58.04 & 46.22 \\
claude-opus-4.5 & 87 & 4 & 1 & 0 & 0 & 6 & 1 & 1 & 188.63 & 134.27 & 748.63 & 59.54 & 55.61 \\
gemini-3-flash-preview & 84 & 0 & 0 & 2 & 2 & 9 & 1 & 2 & 125.04 & 126.09 & 752.86 & 71.03 & 59.18 \\
\rowcolor{agpPurple!8}\quad + Solution-Evo & 98 & 1 & 0 & 0 & 0 & 1 & 0 & 0 & 96.30 & 120.00 & 751.09 & 88.33 & 72.38 \\
\cmidrule{2-14}
\textbf{Gain (\%)} & \textbf{16.7$\uparrow$} & --- & 0 & \textbf{100$\uparrow$} & \textbf{100$\uparrow$} & \textbf{88.9$\uparrow$} & \textbf{100$\uparrow$} & \textbf{100$\uparrow$} & \textbf{23.0$\uparrow$} & \textbf{4.8$\uparrow$} & 0.2$\downarrow$ & \textbf{24.4$\uparrow$} & \textbf{22.3$\uparrow$} \\

\midrule
\rowcolor{agpBlue!7} \multicolumn{14}{c}{\textit{\textbf{Go}}} \\
deepseek-v3.2 & 7 & 0 & 0 & 39 & 0 & 0 & 1 & 53 & 112.71 & 12.59 & 709.57 & 62.73 & 54.36 \\
grok-4.1-fast & 69 & 3 & 0 & 16 & 0 & 4 & 3 & 5 & 194.90 & 23.26 & 755.43 & 66.83 & 62.44 \\
claude-sonnet-4.5 & 44 & 41 & 0 & 0 & 0 & 13 & 0 & 2 & 222.64 & 19.71 & 712.55 & 57.09 & 53.32 \\
claude-opus-4.5 & 84 & 5 & 0 & 0 & 0 & 9 & 0 & 2 & 162.50 & 19.95 & 744.45 & 72.91 & 63.00 \\
gemini-3-flash-preview & 82 & 1 & 0 & 9 & 0 & 7 & 0 & 1 & 139.22 & 22.01 & 739.46 & 76.22 & 63.48 \\
\rowcolor{agpPurple!8}\quad + Solution-Evo & 95 & 0 & 0 & 0 & 0 & 5 & 0 & 0 & 111.64 & 18.35 & 754.17 & 81.52 & 67.94 \\
\cmidrule{2-14}
\textbf{Gain (\%)} & \textbf{15.9$\uparrow$} & \textbf{100$\uparrow$} & 0 & \textbf{100$\uparrow$} & 0 & \textbf{28.6$\uparrow$} & 0 & \textbf{100$\uparrow$} & \textbf{19.8$\uparrow$} & \textbf{16.6$\uparrow$} & \textbf{2.0$\uparrow$} & \textbf{7.0$\uparrow$} & \textbf{7.0$\uparrow$} \\

\midrule
\rowcolor{agpBlue!7} \multicolumn{14}{c}{\textit{\textbf{Kotlin}}} \\
deepseek-v3.2 & 7 & 2 & 0 & 0 & 0 & 0 & 1 & 48 & 59.29 & 62.27 & 793.57 & 58.33 & 74.56 \\
grok-4.1-fast & 62 & 2 & 0 & 22 & 0 & 8 & 2 & 4 & 307.45 & 75.45 & 759.55 & 78.12 & 72.83 \\
claude-sonnet-4.5 & 42 & 36 & 1 & 8 & 1 & 10 & 1 & 1 & 192.62 & 78.49 & 757.64 & 81.59 & 77.79 \\
claude-opus-4.5 & 83 & 4 & 0 & 5 & 0 & 5 & 0 & 3 & 210.47 & 76.60 & 750.98 & 83.18 & 76.53 \\
gemini-3-flash-preview & 75 & 2 & 0 & 8 & 1 & 10 & 2 & 2 & 171.99 & 72.80 & 760.43 & 83.49 & 79.07 \\
\rowcolor{agpPurple!8}\quad + Solution-Evo & 95 & 1 & 0 & 0 & 0 & 4 & 0 & 0 & 122.83 & 77.88 & 749.38 & 83.58 & 67.21 \\
\cmidrule{2-14}
\textbf{Gain (\%)} & \textbf{26.7$\uparrow$} & \textbf{50$\uparrow$} & 0 & \textbf{100$\uparrow$} & \textbf{100$\uparrow$} & \textbf{60$\uparrow$} & \textbf{100$\uparrow$} & \textbf{100$\uparrow$} & \textbf{28.6$\uparrow$} & 7.0$\downarrow$ & 1.5$\downarrow$ & \textbf{0.1$\uparrow$} & 15.0$\downarrow$ \\

\bottomrule
\end{tabular}
\unskip
\end{table}

\textbf{Experimental setup.} AGS uses a planning agent with step bound $m=30$ and five specialist roles, each with $m=20$ and \texttt{gemini-3.1-pro-preview}. These are per-agent limits rather than a cap on total system calls. The \emph{Agent-Evo} variant permits edits to prompts, tools, and code, whereas \emph{Vanilla} retains the initial resources. We report GAIA task-completion accuracy by difficulty and split; HLE uses \texttt{o3-mini} for final judging. Sources are cited in Table~\ref{tab:results2} (see also Appendix~\ref{app:gaia-sources}). The comparison evaluates the complete configured system, including the additional work required to propose and assess edits and execute the updated resources within the stated per-agent step limits.

\textbf{Results.} Agent-Evo increases reported GAIA validation accuracy from 89.70 to 93.33 and test accuracy from 79.07 to 89.04 within the AGS configuration, and AGS scores 59.6 on HLE. Level 3 shows the largest test increase, from 61.22 to 81.63, but no validation increase, indicating that improvements differ across splits and difficulty levels. The within-AGS comparisons share the reported backbone and planner--specialist organization. External leaderboard entries contextualize these scores, but differences in backbones, tools, and inference budgets prevent a controlled comparison.

\subsection{Experiments on Self-Evolving Code Agent Benchmark}

\textbf{Experimental setup.} We compare Vanilla and Solution-Evo using \texttt{gemini-3-flash-preview} on 100 recently released LeetCode problems across five languages. Solution-Evo permits an initial submission and up to two revisions using online-judge feedback. Recent release dates do not establish freedom from training contamination, and other backbones are contextual rather than matched-budget controls. Appendices~\ref{app:coding-benchmark} and~\ref{appx_sec:code-prompts-resources} define the correctness, runtime, memory, and submission-percentile metrics and the associated platform data-use and release policies.

\textbf{Results.} Solution-Evo increases solved counts from 79 to 87 in Python3, 84 to 99 in C++, 84 to 98 in Java, 82 to 95 in Go, and 75 to 95 in Kotlin. These gains characterize execution-guided revision within a task. Mean runtime decreases over each method's accepted solutions, whereas memory and submission-relative percentiles show mixed changes. For example, Kotlin's higher solved count accompanies lower mean runtime but higher mean memory, illustrating why acceptance and execution metrics should be reported separately. Different accepted-problem sets prevent a per-problem efficiency conclusion: this requires paired measurements on commonly solved problems and accounting for extra submissions and reflection calls.

\begin{figure}[!t]
  \centering
  \includegraphics[width=0.9\textwidth,height=0.20\textheight]{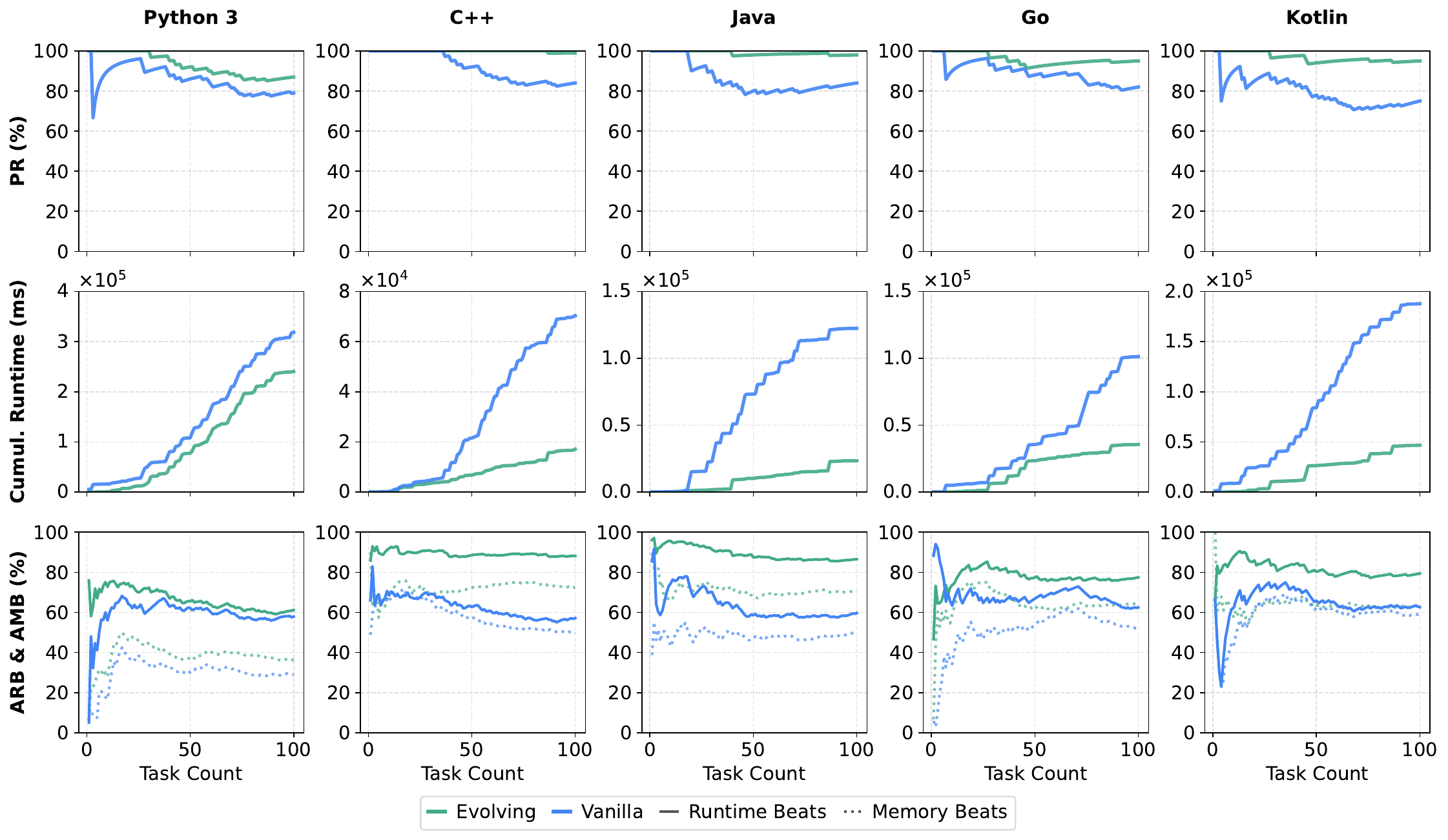}
  \caption{Coding metrics over task count for Vanilla and Solution-Evo.}
  \label{fig:evolution-performance-comparison}
\end{figure}

\textbf{Error profile.} Solution-Evo has no remaining compilation errors in any language. Wrong answers remain the largest failure category in Python3, Go, and Kotlin, so successful compilation alone does not ensure correct solutions. These aggregate categories do not trace individual repairs. Figure~\ref{fig:evolution-performance-comparison} indexes evaluated tasks, not optimization rounds; its curves aggregate outcomes as the evaluated task set grows, rather than tracking successive revisions of a single program.

\subsection{Resource-Update Cases}
\label{sec:resource-cases}

The cases in Appendix~\ref{app:cases} connect resource interfaces to execution decisions. Spreadsheet processing requires a new Tool for reading, a new Agent for auditable aggregation because the built-in implementation is frozen, and a Prompt update separating verification from the answer. A map dead end instead motivates changing the executable Environment. Resource selection thus addresses the source of failure while keeping task inputs and scoring fixed.

The Memory case changes retention and retrieval after compaction. On the same 45-record workload, returned constraints increase from zero to eight, with 25 ordinary recent records preserved. Because the updated policy also adds storage, this is a retrieval result rather than a task-accuracy gain. Across the three cases, different failure signals lead to resource-specific edits under the same sequence of candidate checks, commit, and subsequent invocation. The edited object determines the check: spreadsheet processing checks returned data, the Environment case checks transition behavior, and the Memory case checks retrieved records. Each check concerns the edited resource's behavior; an accepted version becomes available for subsequent task execution through its registered interface.

\section{Limitations and Impact Statement}
\label{sec:limitation}

\textbf{Empirical scope.} The quantitative evaluations assess configured AGS pipelines, including the work required to propose and evaluate edits. GPT-4o controls primary generation opportunities rather than total computation; each AIME set has only 30 questions. The cases illustrate selected accepted updates, not population-level gains. We do not isolate resource management from the update strategy or establish equal-cost superiority, cross-task transfer, or recovery after faulty activation. Conclusions concern the interface and its evaluated uses, not general deployment reliability.

\textbf{Safety and data governance.} Version histories provide inspection records and restoration targets for retained internal state; they do not guarantee behavioral safety or monotonic improvement. Self-evaluation may accept incorrect changes, and external actions may be irreversible. Executable edits require deployment-specific isolation and permissions; traces and memory require access, redaction, and retention controls, separate from the protocol interface. The public release will exclude LeetCode-derived content, task-level traces, credentials, and platform-specific collection/submission tools. Platform use remains subject to the applicable terms; Appendix~\ref{appx_sec:code-prompts-resources} specifies these release exclusions and the resulting limits on reproducing evaluations on the original platform tasks.

\section{Conclusion}

We introduced AGP, a two-layer protocol separating heterogeneous resource management from update strategies. AGS instantiates it through planner--specialist execution; three cases illustrate evaluation, versioned commit, and subsequent use across five resource types. The experiments show task-level gains at additional computational cost, and the cases expose concrete resource changes and activation decisions. AGP thus provides a common lifecycle for heterogeneous resource updates while keeping proposal strategies and resource-specific execution contracts separate.

\clearpage
\bibliography{iclr2027_conference}
\bibliographystyle{iclr2027_conference}

\clearpage
\appendix

\newenvironment{appendixalgorithm}[1]{%
  \par\refstepcounter{algorithm}%
  \begin{tcolorbox}[enhanced jigsaw,breakable,sharp corners,
    colback=white,colframe=black,colbacktitle=white,coltitle=black,
    boxrule=0pt,toprule=0.8pt,bottomrule=0.8pt,titlerule=0.4pt,
    boxsep=0pt,left=0pt,right=0pt,top=3pt,bottom=3pt,
    lefttitle=0pt,righttitle=0pt,toptitle=3pt,bottomtitle=3pt,
    before skip=8pt,after skip=8pt,lines before break=5,fontupper=\normalsize,
    pad at break*=3pt,
    title={\textbf{Algorithm \thealgorithm}\quad #1},
    title after break={\textbf{Algorithm \thealgorithm}\quad #1 (continued)}]%
}{\end{tcolorbox}}

\section{Qualitative Resource-Evolution Cases}
\label{app:cases}

These selected cases use \texttt{google/gemini-3.1-pro-preview} and cover all five resource types. Figures paraphrase task requests and condense execution traces; the Memory policy is shown as a simplified sketch. All shown updates are accepted, so the cases do not exercise version rollback.

\subsection{Tool, Agent, and Prompt: auditable spreadsheet aggregation}
\label{app:case-tool-agent-prompt}

\begin{figure}[H]
\centering
\includegraphics[width=\textwidth]{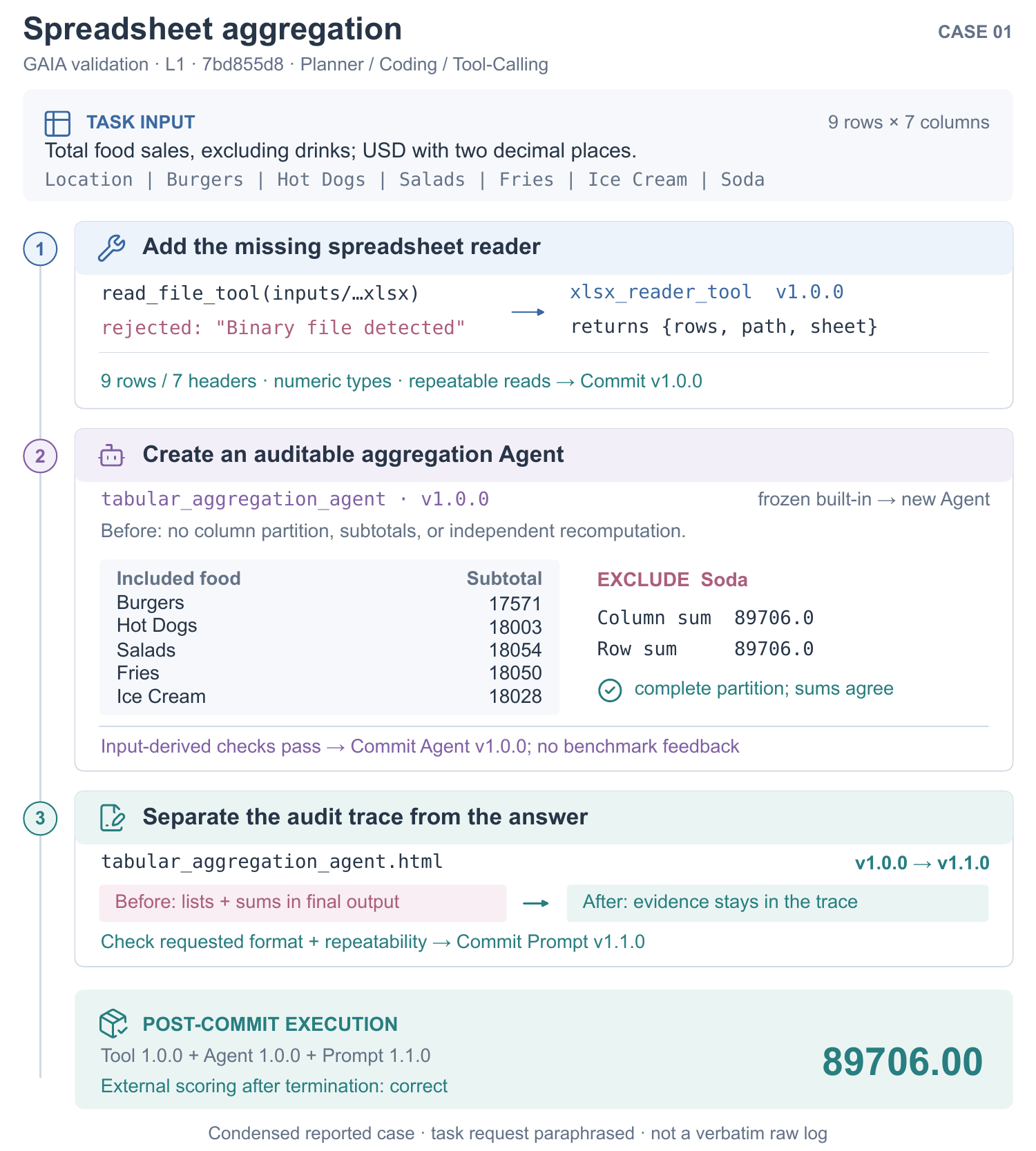}
\caption{Tool, Agent, and Prompt evolution on a GAIA spreadsheet task. Input-derived checks precede each commit; the final benchmark comparison is external to the update loop.}
\label{fig:case-tool-agent-prompt}
\end{figure}

\paragraph{Interpretation.} The frozen built-in Agent is not edited: a new Agent supplies the missing aggregation contract. The later Prompt update is task-independent and contains neither an expected value nor specific columns; formatting follows the user's request.

\FloatBarrier
\subsection{Environment: correcting an executable transition model}
\label{app:case-environment}

\begin{figure}[H]
\centering
\includegraphics[width=\textwidth]{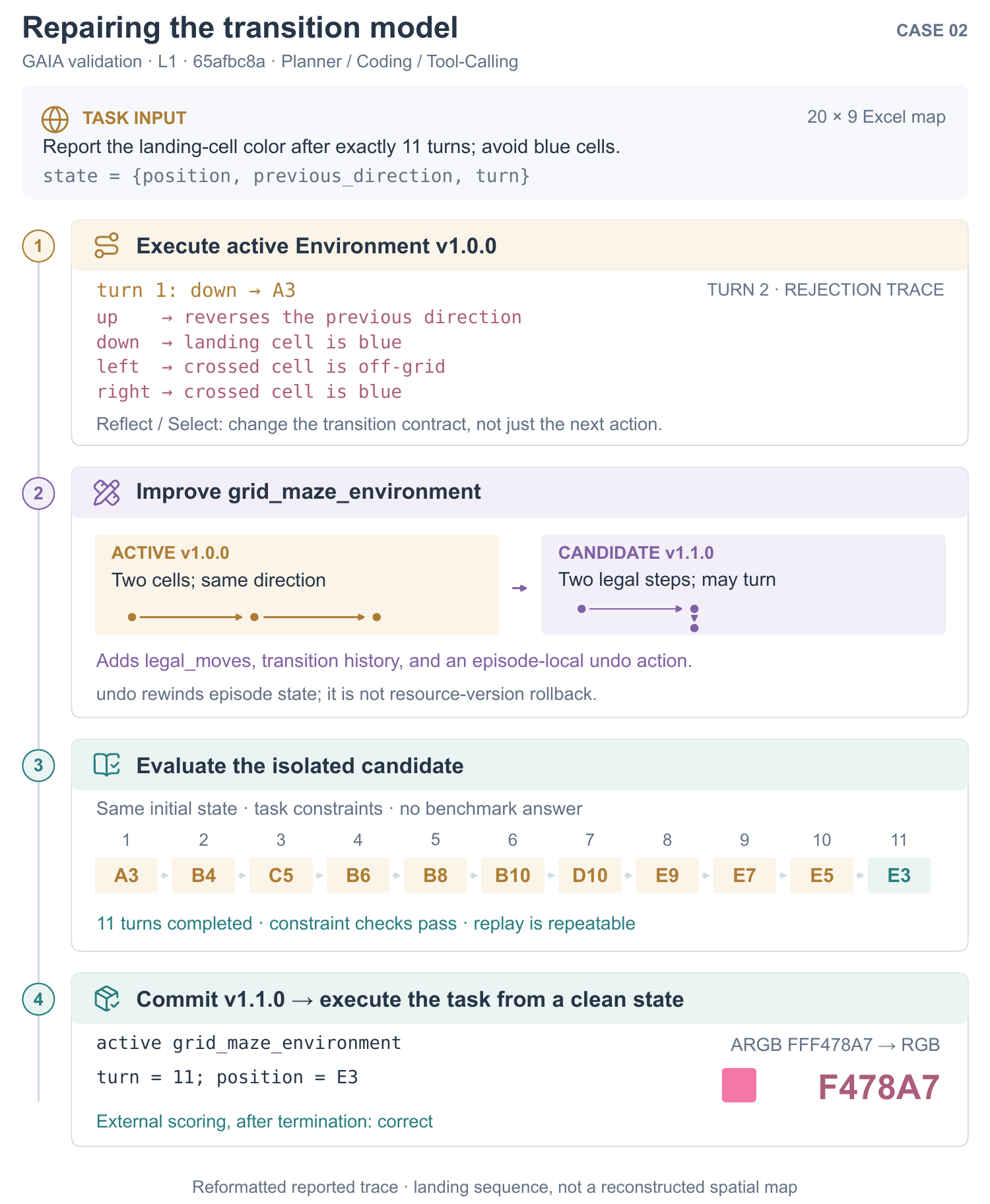}
\caption{Environment evolution on a GAIA map task. The candidate changes executable transition semantics; task rules and map remain fixed inputs. The timeline lists landing cells from the trace, not a reconstructed spatial map.}
\label{fig:case-environment}
\end{figure}

\paragraph{Interpretation.} The planner first registers the generated Environment, whose callable interface passes initial checks but whose transition semantics fail during execution. Candidate evaluation uses the task specification, not an expected terminal cell or benchmark answer. Episode-local \texttt{undo} is distinct from resource-version restoration.

\FloatBarrier
\subsection{Memory: constraint retention through the runtime interface}
\label{app:case-memory}

\begin{figure}[H]
\centering
\includegraphics[width=\textwidth]{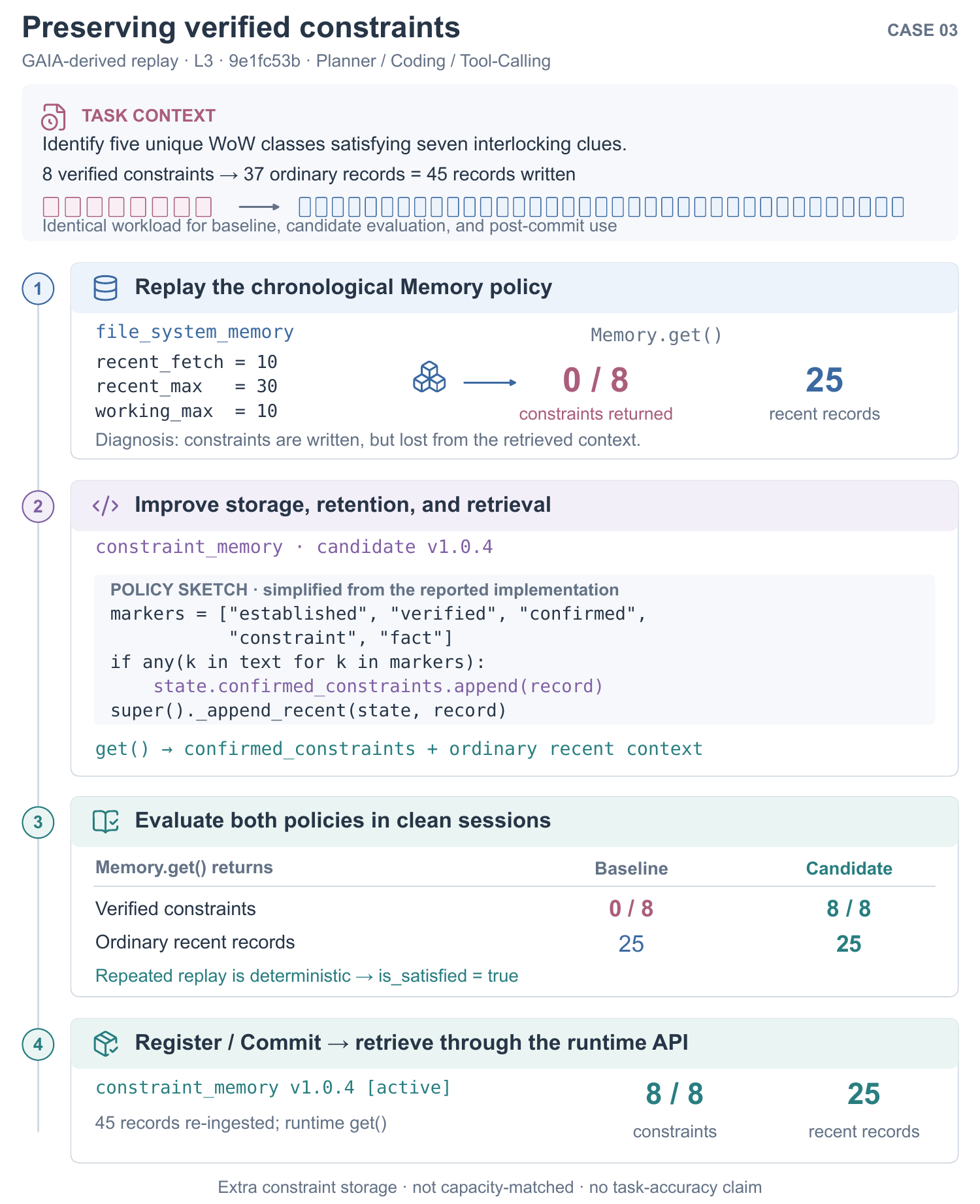}
\caption{Memory-policy replay on a GAIA-derived workload. The log-style panels summarize retrieved records before and after the update; the policy sketch is simplified from the implementation.}
\label{fig:case-memory}
\end{figure}

\paragraph{Interpretation.} The case measures what the active Memory resource returns to the prompt, not final task accuracy. The updated policy adds a persistent collection, so part of the gain may come from the extra storage.

\FloatBarrier

\section{Repeated Evaluations and Feedback Boundaries}
\label{app:repeated-controls}

\paragraph{Configuration and interpretation.}
For the repeated evaluations, GPT-4o is accessed through OpenRouter: \texttt{openrouter/gpt-4o} resolves to \texttt{openai/gpt-4o}. Decoding uses temperature 0.7 and frequency penalty 0.3, with no specified top-$p$ or seed. The reported ceiling is 32,768 output tokens per call. No browser, search, or retrieval tools are exposed. Three complete evaluations use the same benchmark snapshot, source prompts, parser, and method-specific evaluator. Repetition captures pipeline sampling and judge variation, not three disjoint datasets.

\begin{table}[H]
\centering
\caption{Correct/total in each complete GPT-4o evaluation. Each run evaluates 198 GPQA-Diamond questions or 30 AIME questions. SD is the sample standard deviation of three run accuracies, not a confidence interval; each run repeats the same questions.}
\label{tab:repeated-counts}
\small
\setlength{\tabcolsep}{4pt}
\begin{tabular*}{\linewidth}{@{\extracolsep{\fill}}llrrrc@{}}
\toprule
Dataset & Method & Run 1 & Run 2 & Run 3 & Mean $\pm$ SD (\%) \\
\midrule
GPQA-Diamond ($n=198$) & Vanilla & 95/198 & 94/198 & 96/198 & $47.98\pm0.51$ \\
 & Self-Consistency & 101/198 & 100/198 & 102/198 & $51.01\pm0.51$ \\
 & Best-of-3 & 104/198 & 105/198 & 106/198 & $53.03\pm0.51$ \\
 & Simple Reflection & 107/198 & 106/198 & 108/198 & $54.04\pm0.51$ \\
 & PS-Joint-Evo & 115/198 & 118/198 & 118/198 & $59.09\pm0.87$ \\
\midrule
AIME24 ($n=30$) & Vanilla & 4/30 & 3/30 & 5/30 & $13.33\pm3.33$ \\
 & Self-Consistency & 4/30 & 3/30 & 5/30 & $13.33\pm3.33$ \\
 & Best-of-3 & 4/30 & 4/30 & 5/30 & $14.44\pm1.92$ \\
 & Simple Reflection & 3/30 & 2/30 & 4/30 & $10.00\pm3.33$ \\
 & PS-Joint-Evo & 5/30 & 4/30 & 6/30 & $16.67\pm3.33$ \\
\midrule
AIME25 ($n=30$) & Vanilla & 2/30 & 1/30 & 3/30 & $6.67\pm3.33$ \\
 & Self-Consistency & 2/30 & 1/30 & 3/30 & $6.67\pm3.33$ \\
 & Best-of-3 & 2/30 & 3/30 & 3/30 & $8.89\pm1.92$ \\
 & Simple Reflection & 3/30 & 2/30 & 4/30 & $10.00\pm3.33$ \\
 & PS-Joint-Evo & 4/30 & 3/30 & 5/30 & $13.33\pm3.33$ \\
\bottomrule
\end{tabular*}

\end{table}

\begin{table}[H]
\centering
\caption{Three complete GPT-4o evaluations. Each cell reports mean accuracy $\pm$ sample SD (\%), pooled correct/total responses, and approximate output tokens per problem. Pooled denominators are $3\times198=594$ for GPQA-Diamond and $3\times30=90$ for each AIME set, not distinct-question counts. Methods match primary opportunities, not total tokens or calls; Best-of-3 costs include judge output.}
\label{tab:opportunity-controls-detail}
\small
\setlength{\tabcolsep}{8pt}
\renewcommand{\arraystretch}{1.18}
\begin{tabularx}{\linewidth}{@{}l*{3}{>{\centering\arraybackslash}X}@{}}
\toprule
Method & GPQA-Diamond & AIME24 & AIME25 \\
\midrule
Vanilla & \makecell{$47.98\pm0.51$\\285/594\\299 tokens} & \makecell{$13.33\pm3.33$\\12/90\\365 tokens} & \makecell{$6.67\pm3.33$\\6/90\\304 tokens} \\
Self-Consistency & \makecell{$51.01\pm0.51$\\303/594\\896 tokens} & \makecell{$13.33\pm3.33$\\12/90\\1,095 tokens} & \makecell{$6.67\pm3.33$\\6/90\\913 tokens} \\
Best-of-3 & \makecell{$53.03\pm0.51$\\315/594\\1,033 tokens} & \makecell{$14.44\pm1.92$\\13/90\\1,279 tokens} & \makecell{$8.89\pm1.92$\\8/90\\1,075 tokens} \\
Simple Reflection & \makecell{$54.04\pm0.51$\\321/594\\1,141 tokens} & \makecell{$10.00\pm3.33$\\9/90\\1,556 tokens} & \makecell{$10.00\pm3.33$\\9/90\\1,538 tokens} \\
\rowcolor{agpPurple!8}
PS-Joint-Evo & \makecell{$\mathbf{59.09\pm0.87}$\\351/594\\1,838 tokens} & \makecell{$\mathbf{16.67\pm3.33}$\\15/90\\2,005 tokens} & \makecell{$\mathbf{13.33\pm3.33}$\\12/90\\2,345 tokens} \\
\bottomrule
\end{tabularx}

\end{table}

\paragraph{Boxplot conventions.}
Quartiles use linear interpolation across the three run accuracies; whiskers span the observed minimum and maximum. Coincident quartiles or medians reflect repeated outcomes, not missing observations. The AIME panels share a vertical scale, while GPQA-Diamond uses its own scale. Lower bars show output-token costs on a shared zero-based scale and align with the boxes by method. Individual points are omitted for clarity; Table~\ref{tab:repeated-counts} preserves all 45 run counts. With only three runs per method, these summaries do not characterize population distributions.

\paragraph{Question counts and repeated responses.}
The GPQA-Diamond specification gives 198 questions~\citep{rein2024gpqa}; public AIME2024 and AIME2025 releases each contain 30 questions~\citep{h4aime2024,matharenaaime2025}. These sizes agree with the denominators in the reported repeated evaluations. Table~\ref{tab:repeated-counts} reports each run as $k/n$, while Table~\ref{tab:opportunity-controls-detail} sums correct and total responses across three runs. Thus, 594 GPQA-Diamond responses correspond to three evaluations of 198 questions, and 90 AIME responses to three evaluations of 30 questions. These totals do not imply additional unique questions or independent observations across runs.

\subsection{Optimizer-visible information}
\label{app:feedback-boundary}

\begin{table}[H]
\centering
\caption{Feedback policy for GPQA-Diamond/AIME reflection.}
\label{tab:feedback-boundary}
\small
\setlength{\tabcolsep}{4pt}
\begin{tabularx}{\linewidth}{@{}>{\raggedright\arraybackslash}p{0.18\linewidth}>{\raggedright\arraybackslash}p{0.40\linewidth}>{\raggedright\arraybackslash}X@{}}
\toprule
Stage & Available information & Output / restriction \\
\midrule
Reflect / Select / Improve & Question, system instruction, current editable prompt/solution, self-generated reasoning and prior label-free reflection & Proposed change; no reference answer, correctness label, or benchmark score \\
Evaluate & Question and candidate reasoning/answer; a separate call to the same backbone & \texttt{is\_satisfied} (Boolean), \texttt{reasoning} (text); consistency, completeness, and format checks, not exact-match correctness \\
Commit / stop & Candidate identity, edited target, and evaluation result & Record an accepted resource version or retain an accepted task-local solution; stop according to the task configuration \\
Final scoring & Final answer and held-out benchmark answer & Accuracy computed after termination; not returned to the optimizer or stored in an evolving resource \\
\bottomrule
\end{tabularx}
\end{table}

The three-primary-opportunity control means one initial answer and at most two primary revisions for the sequential methods. It is distinct from the generic algorithm's update-iteration budget and from the total number of model calls. A negative self-evaluation can inform the next revision, but neither assessment exposes the reference answer.

Prompt-Evo enables prompt edits while retaining the other resource implementations. Solution-Evo enables edits to the current answer or code without creating a persistent solution resource. PS-Joint-Evo enables both targets within the same task. Here, ``joint'' describes the editable targets, not a separate model-training procedure or evidence of cross-task transfer. Self-evaluation selects or terminates candidates; final benchmark scoring is a separate operation after termination, with no label feedback to the optimizer.

\paragraph{System-level cost.}
AGS uses approximately 49 model-inference steps and 23 minutes per GAIA task, 67 steps and 11 minutes per HLE task (including judging), and 12 steps and 4 minutes per coding task. Total API expenditures are approximately USD 2,000, 6,000, and 200, respectively; the GAIA/HLE experiments ran in April 2026 and the coding experiments in February 2026. These are aggregate figures, not matched-cost comparisons between variants.

\paragraph{Table generation.}
Figure~\ref{fig:opportunity-controls} and Tables~\ref{tab:repeated-counts}--\ref{tab:opportunity-controls-detail} are generated from one machine-readable summary of the GPT-4o counts and output-token costs; means and sample SDs are computed from integer counts before rounding.

Table~\ref{tab:results1} reports one complete evaluation per model and method as accuracy (correct/total), with $n=198$ for GPQA-Diamond and $n=30$ for AIME. Accuracy is $100k/n$ and relative Gain is $100(k_{\rm joint}-k_{\rm vanilla})/k_{\rm vanilla}$, both computed from counts and rounded only for display. These counts belong to the multi-model evaluation and are separate from the repeated GPT-4o controls.

Table~\ref{tab:results3} is likewise generated from a machine-readable coding summary, with Gain computed only for the Gemini Vanilla--Solution-Evo pair in each language; Appendix~\ref{app:coding-benchmark} explains the aggregation scope.

\subsection{GAIA comparison and reporting conventions}
\label{app:gaia-sources}
Table~\ref{tab:results2} cites a source beside each external system: a paper or report where available, otherwise a project page, SDK, or the official GAIA leaderboard results (agent-2030 and Su-Zero-Ultra~\citep{gaia2025results}). The Skywork entry cites the first AgentOrchestra report. External systems use their own backbones, tools, budgets, and evaluation snapshots, so the table provides context rather than a controlled comparison.

Attempt conventions also differ: Alita reports its 87.27 validation result as pass@3~\citep{qiu2025alita}. We list the published value without converting it to pass@1.

The official GAIA evaluator checks split membership and per-level counts, and computes the overall score across questions rather than as an unweighted average of level scores~\citep{gaia2026evaluator}. Our within-AGS scores follow these full-split definitions, and Gain entries are computed from the displayed percentages.

\Needspace{10\baselineskip}
\section{Notation}
\label{appx_sec:notation}

We summarize the main mathematical symbols and their meanings in Table~\ref{appx_tab:notation}.
Symbols follow the two-layer structure of \projectname: resource identities and registration records in RSPL, followed by editable variables, operator messages, and iteration states in SEPL.

\begingroup
  \footnotesize
  \renewcommand{\arraystretch}{0.85}
  \setlength{\tabcolsep}{3pt}
  \begin{longtable}{@{}p{0.24\linewidth}p{\dimexpr0.76\linewidth-2\tabcolsep\relax}@{}}
    \caption{Notation used in the paper. Grey rows indicate categories.}
    \label{appx_tab:notation}\\
    \toprule
    \textbf{Symbol} & \textbf{Description} \\
    \midrule
    \endfirsthead
    \multicolumn{2}{l}{\textit{Table~\thetable{} (continued)}}\\
    \toprule
    \textbf{Symbol} & \textbf{Description} \\
    \midrule
    \endhead
    \midrule
    \multicolumn{2}{r}{\textit{Continued on the next page}}\\
    \endfoot
    \bottomrule
    \endlastfoot
    \rowcolor{gray!15}
    \multicolumn{2}{c}{\textit{\textbf{Indexing and Sets}}} \\
    $\Tau$ & Set of RSPL entity types, $\{\textsc{Prompt},\textsc{Agent},\textsc{Tool},\textsc{Env},\textsc{Mem}\}$. \\
    $\tau$ & Entity type index, $\tau \in \Tau$. \\
    $\mathcal{I}_{\tau}$ & Index set of registered resources of type $\tau$, not temporary runtime objects. \\
    $i$ & Registered-resource index, $i \in \mathcal{I}_{\tau}$. \\
    $\mathbb{V}$ & Space of version strings. \\
    $\wp(\cdot)$ & Power set operator. \\
    \midrule
    \rowcolor{gray!15}
    \multicolumn{2}{c}{\textit{\textbf{RSPL Resource Entity (Def.~\ref{appx_def:resource-entity})}}} \\
    $e_{\tau,i}$ & Resource entity tuple $(n_{\tau,i}, d_{\tau,i}, \phi_{\tau,i}, g_{\tau,i}, m_{\tau,i})$. \\
    $n_{\tau,i}$ & Name unique within type $\tau$; $(\tau,n_{\tau,i})$ is the resource identity. \\
    $d_{\tau,i}$ & Short description. \\
    $\phi_{\tau,i}:\mathcal{X}_{\tau}\!\rightarrow\!\mathcal{Y}_{\tau}$ & Invocation mapping, with runtime state included for stateful resources. \\
    $g_{\tau,i}$ & Evolvability marker, $g_{\tau,i} \in \{0,1\}$, indicating whether the resource is evolvable. \\
    $m_{\tau,i}$ & Auxiliary metadata dictionary. \\
    $\mathcal{E}_{\tau}$ & Set of resource entities of type $\tau$. \\
    \midrule
    \rowcolor{gray!15}
    \multicolumn{2}{c}{\textit{\textbf{RSPL Registration Record (Def.~\ref{appx_def:resource-registration-record})}}} \\
    $c_{\tau,i}$ & Registration record $(e_{\tau,i}, v_{\tau,i}, \eta_{\tau,i}, \theta_{\tau,i}, \mathcal{F}_{\tau,i})$. \\
    $\mathcal{C}_{\tau}$ & Current registration records for type $\tau$; historical records are retained separately. \\
    $v_{\tau,i}$ & Version string; $(\tau,n_{\tau,i},v_{\tau,i})$ identifies a version-qualified record. \\
    $\eta_{\tau,i}$ & Implementation descriptor (e.g., import path, class, or source). \\
    $\theta_{\tau,i}$ & Instantiation parameters (e.g., constructor arguments). \\
    $\mathcal{F}_{\tau,i}$ & Exported representations for LLM interaction (schemas/text/structured args). \\
    \midrule
    \rowcolor{gray!15}
    \multicolumn{2}{c}{\textit{\textbf{Type-level Resource Interface (Def.~\ref{appx_def:protocol-registered-resource})}}} \\
    $\mathcal{R}_{\tau}$ & Type-specific registry mapping identities to active registration records. \\
    $\mathcal{R}$ & Global registry, $\bigcup_{\tau}\mathcal{R}_{\tau}$, with type-qualified keys. \\
    $\mathcal{M}_{\tau}$ & Context manager for type $\tau$ (maintains registry and version lineage). \\
    $\mathcal{A}_{\tau}$ & Server-exposed interface for type $\tau$ (delegates to $\mathcal{M}_{\tau}$). \\
    $r_{\tau}$ & Type-level management interface $(\mathcal{C}_{\tau}, \mathcal{M}_{\tau}, \mathcal{A}_{\tau})$, not a single instance. \\
    \midrule
    \rowcolor{gray!15}
    \multicolumn{2}{c}{\textit{\textbf{SEPL Variables, Spaces, and Operators}}} \\
    $\mathcal{V}_{\text{evo}}$ & Managed resource entities together with the current task-local output $y$. \\
    $\mathfrak{V}$ & Space of admissible resource/output states, including newly registered instances. \\
    $v$ & A variable in $\mathcal{V}_{\text{evo}}$. \\
    $g_{v}$ & Evolvability marker for variable $v$ (binary). \\
    $\Theta$ & Editable subset, $\{v \in \mathcal{V}_{\text{evo}} \mid g_{v}=1\}$. \\
    $y$ & Current task-local output or solution; not an editable execution trace. \\
    $\mathcal{P}$ & Message space carrying evidence, provisional candidates, and assessments between operators. \\
    $\mathcal{P}_{\text{in}}, \mathcal{P}_{\text{out}}$ & Incoming and outgoing message types of a SEPL operator. \\
    $f$ & A SEPL operator, $f: \mathfrak{V} \times \mathcal{P}_{\text{in}} \rightarrow \mathfrak{V} \times \mathcal{P}_{\text{out}}$. \\
    $\mathcal{Z}$ & Trace space (execution observations). \\
    $\mathcal{H}$ & Hypothesis space (proposed explanations of observed failures). \\
    $\mathcal{D}$ & Modification space (proposed resource changes). \\
    $\mathcal{G}$ & Objective specification (task goals and declared evaluation constraints). \\
    $\mathcal{S}$ & Evaluation space (reported metrics and constraint-check outcomes). \\
    $\rho,\sigma,\iota,\varepsilon,\kappa$ & Reflect, Select, Improve, Evaluate, and Commit operators (reflection instantiation). \\
    \midrule
    \rowcolor{gray!15}
    \multicolumn{2}{c}{\textit{\textbf{Optimization Loop (Alg.~\ref{alg:sepl-loop-appx})}}} \\
    $A$ & Agentic system. \\
    $T$ & Optimization budget (number of iterations). \\
    $t$ & Iteration index. \\
    $\mathcal{V}_{\text{evo}}^{(t)}$ & Evolvable state at iteration $t$. \\
    $p^{(t)}\in\mathcal{P}$ & Message instance at iteration $t$; the initial message carries an execution trace. \\
    $Z^{(t)}\in\mathcal{Z}$ & Observational trace at iteration $t$. \\
    $H^{(t)}\subseteq\mathcal{H}$ & Hypotheses at iteration $t$. \\
    $D^{(t)}\subseteq\mathcal{D}$ & Proposed modifications at iteration $t$. \\
    $\widetilde{\mathcal{V}}_{\text{evo}}^{(t+1)}$ & Candidate state after applying modifications. \\
    $S^{(t+1)}\in\mathcal{S}$ & Evaluation result for the candidate state. \\
    $F\in\mathcal{S}^{*}$ & Ordered candidate-assessment history supplied to reflection; distinct from execution trace $Z$. \\
  \end{longtable}
\endgroup

\section{Code, Prompts, and Resources}
\label{appx_sec:code-prompts-resources}

\paragraph{Reproducibility materials.}
Appendix~\ref{app:repeated-controls} records the repeated-run counts and configuration; Appendix~\ref{app:feedback-boundary} specifies optimizer-visible fields. Machine-readable summaries reproduce the reasoning and coding tables and the repeated-run figure. The case figures condense selected execution traces.

\paragraph{LeetCode-derived materials.}
The public release will contain the generic framework, aggregate experimental settings and results, and methodological descriptions, but no LeetCode-derived problem statements, examples, constraints, starter code, test cases, reference solutions, problem identifiers, fixed splits, or per-problem derived metadata. We will also exclude task-level prompts and traces, platform-specific collection/submission tools, and account credentials. These exclusions limit independent reproduction on the original problem set.

\paragraph{Platform use and data handling.}
Platform-facing access and evaluation must comply with the applicable LeetCode terms and access conditions~\citep{leetcode2024terms}. The experiments use only judge verdicts and execution metadata; hidden tests and official reference solutions are never accessed. Model inference runs through third-party APIs, so traces and requests are subject to the providers' data-handling policies.

\subsection{Implementation details}
\label{app:implementation-evidence}

\paragraph{Resource management.}
Prompt resources expose names, versions, evolvability flags, templates, and rendering variables; the prompt server delegates lifecycle operations to its context manager. The version manager stores version metadata, while retained prompt configurations reside in the prompt context manager. Unregister removes the active prompt binding but retains its local history; cleanup clears that manager's configurations and history. Restoration acts on retained prompt state and is not a system-wide transaction.

\paragraph{Datasets.}
The reasoning benchmarks contain 198 GPQA-Diamond questions, 30 AIME24 problems, and 30 AIME25 problems (released in two 15-problem parts). The GPQA-Diamond loader shuffles answer options with a fixed preprocessing seed of 42; this is not an LLM decoding seed.

\paragraph{State and feedback boundaries.}
Benchmark \texttt{reset} methods reset the item index and collected results, not every resource registry, memory store, or cache. Task records store the reference answer separately from the question, and label-free callers pass only the fields listed in Table~\ref{tab:feedback-boundary} to the optimizer.

\section{Comparison with Other Protocols}
\label{appx_sec:comparison-with-other-protocols}

MCP and A2A address communication and interoperability concerns complementary to AGP~\citep{mcp2025spec,a2a2026spec}. Table~\ref{appx_tab:protocol_comparison} compares their primary management objects and lifecycle boundaries. A client--server connection, an inter-agent task, and a versioned resource implementation have distinct state transitions, allowing the protocols to be combined within a larger application.

\begin{table}[H]
\centering
\caption{Scope comparison. Application-layer extensions can combine these mechanisms; an item outside a standard's primary scope is not a claim that implementations cannot support it.}
\label{appx_tab:protocol_comparison}
\small
\renewcommand{\arraystretch}{1.2}
\setlength{\tabcolsep}{3pt}
\begin{tabularx}{\linewidth}{@{}>{\raggedright\arraybackslash}p{0.16\linewidth}*{3}{>{\raggedright\arraybackslash}X}@{}}
\toprule
Concern & AGP & MCP & A2A \\
\midrule
Primary object & Editable agent resources and candidate versions & Tools, prompts, and contextual resources & Agents, messages, tasks, and artifacts \\
Lifecycle focus & Resource registration, evaluation, commit, and retained versions & Client--server sessions and capability interaction & Task states and agent interaction \\
Evolution focus & Common interface for proposing and recording resource changes & Resource-edit history is an application concern & Internal implementation updates are an application concern \\
Composition & Can manage a tool or agent exposed through another interface & Can expose capabilities used by AGS & Can connect participants in a larger agent system \\
Evidence limit & Cases show accepted updates; recovery needs testing & Comparison concerns the cited specification & Comparison concerns the cited specification \\
\bottomrule
\end{tabularx}
\end{table}

AGP focuses on associating resource changes with execution evidence and retained versions; this does not imply that other standards or their implementations lack logging, security, or state management. The comparison concerns the cited interfaces rather than implementation-level performance.

\subsection{Comparison with Resource and Workflow Optimizers}
\label{app:optimization-comparison}

Table~\ref{tab:optimization-comparison} compares the mechanisms described in the cited papers. It separates what changes, the scope and feedback of adaptation, and the state retained by the method. Several of these methods already retain search history, episodic memory, or reusable tools.

\begingroup
\footnotesize
\setlength{\tabcolsep}{3pt}
\renewcommand{\arraystretch}{1.12}
\begin{longtable}{@{}>{\raggedright\arraybackslash}p{0.15\linewidth}>{\raggedright\arraybackslash}p{0.20\linewidth}>{\raggedright\arraybackslash}p{0.36\linewidth}>{\raggedright\arraybackslash}p{\dimexpr0.29\linewidth-6\tabcolsep\relax}@{}}
\caption{Optimization mechanisms and retained state. This is a methodological comparison, not a matched-budget ranking.}
\label{tab:optimization-comparison}\\
\toprule
\textbf{Method} & \textbf{Update target} & \textbf{Scope and feedback} & \textbf{Retained state} \\
\midrule
\endfirsthead
\multicolumn{4}{l}{\textit{Table~\thetable{} (continued)}}\\
\toprule
\textbf{Method} & \textbf{Update target} & \textbf{Scope and feedback} & \textbf{Retained state} \\
\midrule
\endhead
\midrule
\multicolumn{4}{r}{\textit{Continued on the next page}}\\
\endfoot
\bottomrule
\endlastfoot
Self-Refine\par\citep{madaan2023self} & Current response & Within-instance revisions using feedback from the same LLM. & Response and feedback history in context. \\
\addlinespace[3pt]
Reflexion\par\citep{shinn2023reflexion} & Reflections guiding later actions & Subsequent trials use verbal reflections on scalar or textual task feedback. & Episodic reflection memory. \\
\addlinespace[3pt]
TextGrad\par\citep{yuksekgonul2025optimizing} & Textual variables in a computation graph & Instance-level solution edits or task-level prompt optimization using backpropagated LLM feedback. & Variable values, graph dependencies, and textual feedback. \\
\addlinespace[3pt]
MIPRO\par\citep{opsahlong2024mipro} & Module instructions and demonstrations & Program optimization over training examples using downstream metrics and Bayesian search. & Candidate combinations and a learned surrogate objective. \\
\addlinespace[3pt]
GEPA\par\citep{agrawal2026gepa} & Module prompts; also code-search examples & Trajectory reflection and task evaluation guide prompt search; the paper also studies inference-time search. & Candidate pool, ancestry, and per-instance scores for Pareto selection. \\
\addlinespace[3pt]
AFlow\par\citep{zhang2025aflow} & Code-represented workflows & Workflow search using validation executions, scores, and feedback on code edits. & Search-tree experience, parent edits, and evaluation records. \\
\addlinespace[3pt]
EvoAgentX\par\citep{wang2025evoagentx} & Prompts, tool configurations, and workflow topology & Workflow evaluation and optimization through integrated TextGrad, MIPRO, and AFlow components. & Workflow representations and optimizer-specific state. \\
\addlinespace[3pt]
Alita\par\citep{qiu2025alita} & Tools and execution environments & Runtime tool construction and self-correction from execution feedback; generated tools can be reused. & An MCP tool registry and execution logs. \\
\midrule
\rowcolor{agpBlue!7}
AGP / AGS\par(this work) & Registered resources and task-local solutions & Within-task reflection and selected resource updates under declared feedback policies. & Resource identities, accepted versions, and trace-linked decisions. \\
\end{longtable}
\endgroup

AGP specifies a common resource-update interface, whereas the cited optimizers define proposal or search procedures and the state needed to support them. The distinction is not absolute: for example, GEPA records candidate ancestry and AFlow records parent edits, while Alita explicitly stores generated tools for reuse. An AGP adapter would need to map such state to resource identities, candidate evaluation, and activation decisions; a shared interface alone does not verify that mapping. Our main experiments use reflection, and the alternative adapters remain unevaluated sketches.

AGP's management contract requires rejection to preserve the active binding, while restoration targets a retained internal version. The cases document accepted updates, not recovery after a bad version becomes active. The reported target ablations compare what is revised within AGS; they do not isolate resource management while holding the editor, feedback, and realized budget fixed. Persistent storage alone likewise does not establish transfer to unseen tasks.

\FloatBarrier

\section{Details of the Self-Evolving Code Agent Benchmark}
\label{app:coding-benchmark}

\subsection{Benchmark Design Rationale}

The coding evaluation studies feedback-driven revision within a single problem-solving episode. It combines three perspectives: whether the final submission is accepted, how accepted solutions use runtime and memory, and how outcomes vary across programming languages. Online-judge feedback supports iterative revision without exposing hidden test cases or reference solutions to the agent.

Compilation errors, runtime failures, wrong answers, and resource-limit violations provide different signals for revision. The agent can use these signals to propose changes to syntax, interfaces, logic, or algorithmic structure. We evaluate final submissions under a declared reflection budget. Figure~\ref{fig:evolution-performance-comparison} presents metrics over task count, rather than successive revisions of one solution.

LeetCode also reports runtime and memory statistics for accepted submissions, including platform-relative percentiles. We use these statistics to describe solution efficiency alongside acceptance counts. They are relative to the platform's submission distributions, which may change over time, and should not be interpreted as measurements against a controlled sample of individual programmers.

The same selected problems are instantiated in Python3, C++, Java, Go, and Kotlin using the corresponding starter interfaces. This setup exposes language-specific compilation, typing, library, and runtime failures while holding the task specifications fixed. Cross-language results therefore characterize the interaction between code generation, execution feedback, and language-specific constraints, rather than treating success in one language as evidence of universal robustness.

\subsection{Benchmark Construction}

\textbf{Data Collection.}
We collected a pool of 3{,}822 LeetCode problems from publicly visible pages, with statements, public examples, constraints, difficulty labels, topical tags, and starter interfaces. Figure~\ref{fig:leetcode-tags-distribution} summarizes the topics of a 3{,}704-record metadata snapshot covering 72 topics. Problems may have multiple tags; the chart counts tag assignments and pools the 52 less frequent topics.

\newlength{\codingpanelheight}
\begin{figure}[!htb]
    \centering
    \setlength{\codingpanelheight}{0.457833\textwidth}
    \begin{minipage}[t]{0.67\textwidth}
        \centering
        \includegraphics[height=\codingpanelheight]{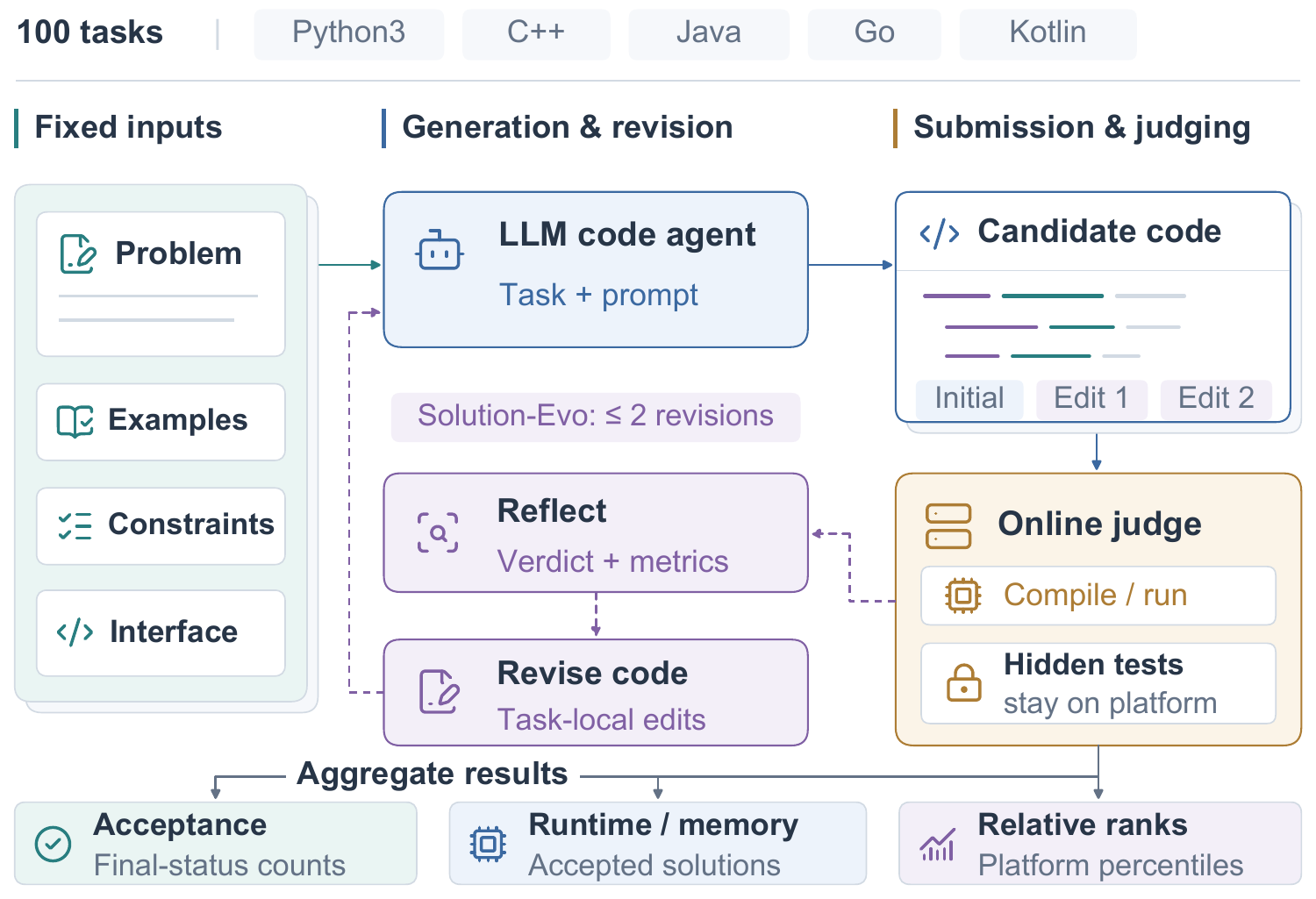}
        \caption{Coding evaluation with feedback-driven revision.}
        \label{fig:code-evaluation-pipeline}
    \end{minipage}\hfill
    \begin{minipage}[t]{0.31\textwidth}
        \centering
        \includegraphics[height=\codingpanelheight]{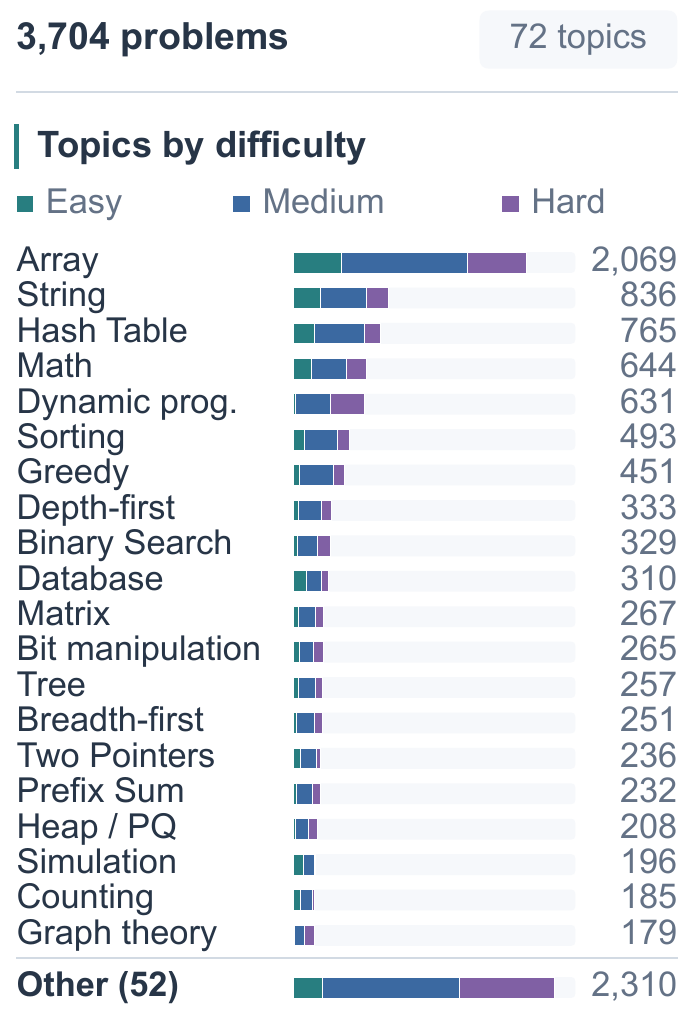}
        \captionsetup{justification=centering}
        \caption{Topic coverage.}
        \label{fig:leetcode-tags-distribution}
    \end{minipage}
\end{figure}

Each collected record is normalized into a task specification, official examples, a language-specific starter template, and difficulty and topic metadata. The preprocessing preserves judge-required signatures and class definitions, filters malformed and duplicate records, and checks template availability for the target languages. We select 100 recently released problems from the collected pool and instantiate them across all five languages. This selection aims to reduce contamination risk, but release recency alone cannot establish whether a task was absent from a model's training data.

\textbf{Problem Characteristics.}
Each task specifies input and output behavior, constraints, and a required submission interface. The selected categories span different algorithmic and data-structure requirements, while hidden tests can expose boundary cases and implementation errors. The common task specifications support comparisons across languages; they do not eliminate language-specific compiler or runtime effects.

\textbf{Problem Evaluation.}
For each problem, the agent receives the task representation and submits code to the platform judge, which checks hidden tests and reports the available execution statistics. The reported Solution-Evo budget permits at most three submissions: the initial solution and up to two feedback-driven revisions, with the task specification and submission interface held fixed. Vanilla uses the initial solution without these revisions. This comparison measures the combined effect of feedback-driven revision and additional attempts, rather than an equal-computation advantage.

\subsection{Evaluation Metrics}

\begin{table}[!htb]
  \centering
  \footnotesize
  \renewcommand{\arraystretch}{0.85}
  \setlength{\tabcolsep}{4pt}
  \caption{Evaluation metrics for the algorithmic coding benchmark.}
  \label{tab:coding-metrics}
  \begin{tabularx}{\linewidth}{@{}p{0.12\linewidth}X@{}}
  \toprule
  \textbf{Metric} & \textbf{Description} \\
  \midrule
  \rowcolor{gray!15} \multicolumn{2}{c}{\textbf{Capability metrics}} \\
  \textbf{PR}  & Number of problems passing all hidden test cases within time and memory limits. \\
  \textbf{TLE} & Number of problems exceeding the allowed execution time limit. \\
  \textbf{MLE} & Number of problems exceeding the allowed memory usage. \\
  \textbf{CE}  & Number of problems where generated code failed to compile. \\
  \textbf{RE}  & Number of problems encountering a runtime error during execution. \\
  \textbf{WA}  & Number of problems producing incorrect output. \\
  \textbf{TO}  & Number of problems where the model failed to respond within the timeout. \\
  \textbf{RpE} & Number of problems where the model returned an invalid or unparseable response. \\
  \midrule
  \rowcolor{gray!15} \multicolumn{2}{c}{\textbf{Execution metrics}} \\
  \textbf{AR}  & Mean runtime in milliseconds over accepted solutions. \\
  \textbf{AM}  & Mean memory usage in megabytes over accepted solutions. \\
  \textbf{APC} & Mean number of test cases passed before failure. \\
  \midrule
  \rowcolor{gray!15} \multicolumn{2}{c}{\textbf{Submission-relative metrics}} \\
  \textbf{ARB} & Mean platform-reported runtime-beats percentile over accepted solutions. \\
  \textbf{AMB} & Mean platform-reported memory-beats percentile over accepted solutions. \\
  \bottomrule
  \end{tabularx}
\end{table}

We report three groups of metrics that capture complementary aspects of code-agent performance. Capability metrics evaluate functional correctness and diagnose execution-blocking failure modes. PR measures the number of fully accepted problems, while TLE, MLE, CE, RE, WA, TO, and RpE identify whether failures arise from algorithmic inefficiency, excessive memory use, compilation errors, runtime exceptions, incorrect logic, response timeout, or invalid output formatting. These metrics are particularly important for self-evolving agents because different failure modes correspond to different refinement opportunities.

AR and AM summarize runtime and memory over each method's accepted solutions. Because those sets can differ, these aggregates do not isolate per-problem efficiency; a common-solved-set paired analysis is required. APC measures passed-test coverage rather than resource efficiency, with larger values denoting more passed cases in the reported aggregate. Its interpretation also depends on which submissions contribute and how many tests each task contains.

Platform-relative metrics aggregate the reported runtime and memory percentiles of accepted submissions. These percentiles describe a changing submission distribution rather than a controlled human baseline. Their interpretation depends on the accepted set and aggregation procedure.

Table~\ref{tab:results3} reports aggregate coding metrics. In Figure~\ref{fig:evolution-performance-comparison}, the horizontal axis is task count; the panels show acceptance rate, cumulative runtime, and submission-relative percentiles. These are not per-round learning curves or evidence of cross-task transfer, and their shape depends on task order.

\paragraph{Gain conventions and aggregation scope.}
For a nonzero Vanilla value $b$ and Solution-Evo value $e$, Table~\ref{tab:results3} reports $100|e-b|/b$. Arrows denote the direction of benefit: increases are favorable for PR, APC, ARB, and AMB, whereas decreases are favorable for error counts, AR, and AM. Zero-to-zero entries are shown as no change; a zero-to-nonzero transition has no defined relative percentage. In particular, Java TLE increases from zero to one and is marked with a dash, not a zero gain. Values displayed as 0.0 indicate changes that round to zero at the stated precision.

Status categories are reported as recorded and need not sum to the number of evaluated problems, so we do not normalize them into error distributions. Gain compares corresponding metrics within each same-backbone Vanilla--Solution-Evo pair, not necessarily identical submissions or accepted sets.

\FloatBarrier
\section{Details of Self-Evolution Protocol}

\subsection{Design Motivation}

The design separates update proposals from resource management. An optimizer needs an identified target, a description of permitted edits, candidate evaluation, and a recorded activation decision. RSPL represents the target and retained versions; SEPL organizes those operations into an update loop. This separation permits common bookkeeping across resource types, while useful recovery additionally depends on which state is captured, which consumers bind to a version, and whether external effects occur.

\subsection{Layer 1: Resource Substrate Protocol Layer}

RSPL represents instructions (Prompt), decision logic (Agent), callable capabilities (Tool), executable environment models (Environment), and retention/retrieval implementations (Memory). Tool includes local code, MCP-connected capabilities~\citep{mcp2025spec}, and skills~\citep{anthropic2025agentskills}. The resource representation does not prescribe its own update strategy. Execution changes task-local state, whereas resource updates revise stored implementations or configurations.

\subsubsection{Core Entities}

We focus on these five entity types as the supported resource substrate for agentic systems. This choice is not intended to be exhaustive, but rather to identify a common denominator across modern agent stacks and provide a uniform target space on which SEPL can operate.

\begin{definition}[Resource Entity]
\label{appx_def:resource-entity}
For type $\tau$, the entity and its collection are
\begin{equation}
\begin{aligned}
e_{\tau,i} &= (n_{\tau,i},\, d_{\tau,i},\, \phi_{\tau,i},\, g_{\tau,i},\, m_{\tau,i}),\\
\mathcal{E}_{\tau} &= \{\, e_{\tau,i} \mid i \in \mathcal{I}_{\tau} \,\},
\end{aligned}
\end{equation}
where $\tau \in \Tau = \{\textsc{Prompt},\textsc{Agent},\textsc{Tool},\textsc{Env},\textsc{Mem}\}$ indexes resource types and $i \in \mathcal{I}_\tau$ indexes registered resources. Each entity has a name $n_{\tau,i}$ unique within its type, description $d_{\tau,i}$, input-to-output mapping $\phi_{\tau,i}:\mathcal{X}_{\tau} \rightarrow \mathcal{Y}_{\tau}$, evolvability marker $g_{\tau,i}\in\{0,1\}$, and auxiliary metadata $m_{\tau,i}$. For stateful resources, invocation inputs and outputs include relevant runtime state; the mapping does not assume pure or stateless execution.
\end{definition}

Representing prompts, tools, and memory as explicit resources separates agent control logic from the instructions and capabilities it consumes. A tool-calling agent can then retain its implementation while selecting different prompt or tool resources. The intended benefit is independent management and reuse of these components; whether an accepted resource improves performance on a new task remains an empirical question.

To support resource registration, unified management, and instantiation, RSPL stores a serializable registration record for each resource instance.
\begin{definition}[Resource Registration Record]
  \label{appx_def:resource-registration-record}
  A resource registration record and its type-level collection can be represented as:
  \begin{equation}
  \begin{aligned}
  c_{\tau,i} &= (e_{\tau,i},\, v_{\tau,i},\, \eta_{\tau,i},\, \theta_{\tau,i},\, \mathcal{F}_{\tau,i}),\\
  \mathcal{C}_{\tau} &= \{\, c_{\tau,i} \mid i \in \mathcal{I}_{\tau} \,\},
  \end{aligned}
  \end{equation}
  where $\tau \in \Tau$ indexes the entity type and $i \in \mathcal{I}_\tau$ indexes an individual instance. Here $e_{\tau,i}$ is the resource entity tuple defined in \cref{appx_def:resource-entity}, $v_{\tau,i} \in \mathbb{V}$ is a version string, $\eta_{\tau,i}$ is an implementation descriptor (e.g., import path, class definition, or source-code string), $\theta_{\tau,i}$ are instantiation parameters (e.g., constructor arguments), and $\mathcal{F}_{\tau,i}$ is a set of exported representations used by LLMs to interact with the resource (e.g., function-calling schema, plain text, and structured argument schema).
  \end{definition}

\begin{definition}[Type-level resource interface]
\label{appx_def:protocol-registered-resource}
For each entity type $\tau$, the registry $\mathcal{R}_\tau$ maps typed identities $(\tau,n_{\tau,i})$ to active records in $\mathcal{C}_\tau$, while historical records remain separate restoration targets. The global registry $\mathcal{R}=\bigcup_\tau\mathcal{R}_\tau$ uses type-qualified keys. RSPL binds each type to a context manager $\mathcal{M}_\tau$ and server interface $\mathcal{A}_\tau$. The type-level management interface, not an individual instance, is
\begin{equation}
r_{\tau} = (\mathcal{C}_\tau,\; \mathcal{M}_\tau,\; \mathcal{A}_\tau),
\end{equation}
where each $c_{\tau,i} \in \mathcal{C}_\tau$ is a registration record in \cref{appx_def:resource-registration-record}. The context manager $\mathcal{M}_\tau$ maintains the collection $\mathcal{C}_\tau$, the version lineage for type $\tau$, and implements lifecycle and update operations over these records; the server-exposed interface $\mathcal{A}_\tau$ encapsulates $\mathcal{M}_\tau$ and exposes a unified external interface by delegating requests to the corresponding context-manager routines.
\end{definition}

\paragraph{Management interface versus invocation contract.}
The triple $r_\tau$ describes management at the resource-type level; $\phi_{\tau,i}$ describes invocation of an individual resource. A shared lifecycle does not require all tools or agents to accept the same arguments. The spaces $\mathcal{X}_\tau$ and $\mathcal{Y}_\tau$ collect the admissible type-level inputs and outputs, while each resource's exported representations $\mathcal{F}_{\tau,i}$ specify its own accepted schema. Consumers must resolve that contract before invocation. Updating an implementation without changing its callable schema preserves this part of the consumer interface; changing the schema may require corresponding consumer changes.

\paragraph{Identity, version, and runtime instance.}
The pair $(\tau,n_{\tau,i})$ identifies a registered resource; $(\tau,n_{\tau,i},v_{\tau,i})$ identifies a version-qualified record. Here $c_{\tau,i}$ denotes the current record, while retained historical records belong to the manager's history rather than additional entries in the active registry. A version update preserves the resource identity. Constructing another runtime object from a record also does not create a new registered resource; registration under a different typed name does. Typed names distinguish identically named resources across separate registries.

\subsubsection{Context Manager}
\label{appx_sec:context-manager}

The context manager maintains the active registry and retained version history. Table~\ref{appx_tab:context-manager-interface} defines the protocol's operation vocabulary, not a list of identically named Python methods implemented by every resource type. An adapter maps supported operations to a type's native interface and declares unavailable capabilities. Exported schemas and text describe resource interaction, including local tools, MCP-connected capabilities~\citep{mcp2025spec}, and skills~\citep{anthropic2025agentskills}. Such descriptions inform consumers about arguments and intended use; they do not certify correctness or enforce the stated restrictions.

\begin{table}[!ht]
\centering
\small
\caption{Conceptual RSPL management operations. Native names and availability vary by type; optional operations require explicit adapter support.}
\label{appx_tab:context-manager-interface}
\begin{tabularx}{\linewidth}{@{}p{0.21\linewidth}X@{}}
\toprule
\textbf{Operator} & \textbf{Description} \\
\midrule
\rowcolor{gray!15} \multicolumn{2}{c}{\textbf{Lifecycle \& Registration}} \\
$\mathtt{init}$ & Discover and initialize configured resources; exposed as \texttt{initialize} in AGS. \\
$\mathtt{build}$ & Construct a runtime object from its record; not a separate method on every server. \\
$\mathtt{register}$ & Add a resource under a typed name with an associated version. \\
$\mathtt{unregister}$ & Remove the active registration. History retention is manager-specific; deletion requires a separate policy, and deleted history cannot be restored. \\
\midrule
\rowcolor{gray!15} \multicolumn{2}{c}{\textbf{Retrieval \& Inspection}} \\
$\mathtt{get}$ & Retrieve a resource instance by name from the active registry. \\
$\mathtt{get\_info}$ & Retrieve a resource configuration by name from the active registry. \\
$\mathtt{list}$ & List all registered resource names. \\
$\mathtt{retrieve}$ & Optional similarity-based resource lookup; not natively implemented in AGS. \\
$\mathtt{get\_state}$ & Get the current state of a resource instance when supported. \\
\midrule
\rowcolor{gray!15} \multicolumn{2}{c}{\textbf{Versioning}} \\
$\mathtt{update}$ & Request an implementation/configuration change under the manager's version policy. \\
$\mathtt{copy}$ & Duplicate a resource with an optional new name and version. \\
$\mathtt{restore}$ & Restore a specific historical version by name and version string. \\
$\mathtt{get\_variables}$ & Adapter-level projection of permitted fields from a registration record. \\
$\mathtt{set\_variables}$ & Adapter-level application of a candidate projection, subject to evaluation and commit. \\
\midrule
\rowcolor{gray!15} \multicolumn{2}{c}{\textbf{Execution}} \\
$\mathtt{run}$ & Invoke the type-specific execution surface; not a shared native method signature. \\
\midrule
\rowcolor{gray!15} \multicolumn{2}{c}{\textbf{Serialization}} \\
$\mathtt{save\_to\_json}$ & Optional export of configurations/history to a declared persistence format. \\
$\mathtt{load\_from\_json}$ & Optional import with type, version, and artifact checks supplied by the adapter. \\
$\mathtt{save\_contract}$ & Optional persistence of the exported interaction description. \\
$\mathtt{load\_contract}$ & Optional loading of that description, not activation of its implementation. \\
\bottomrule
\end{tabularx}
\end{table}

\subsubsection{Server Interface}
\label{appx_sec:server-interface}

In AGS, the server classes are local Python facades over context managers rather than a network transport or wire protocol. All five directly declare \texttt{initialize}, \texttt{register}, \texttt{get}, \texttt{get\_info}, \texttt{list}, \texttt{update}, \texttt{unregister}, and \texttt{cleanup}. Their argument and return types vary: this is not a uniform server API. Table~\ref{tab:native-resource-interfaces} maps representative execution and version operations to these declarations.

\begin{table}[!ht]
\centering
\small
\caption{Native resource interfaces in AGS.}
\label{tab:native-resource-interfaces}
\setlength{\tabcolsep}{4pt}
\begin{tabularx}{\linewidth}{@{}>{\raggedright\arraybackslash}p{0.15\linewidth}>{\raggedright\arraybackslash}p{0.43\linewidth}>{\raggedright\arraybackslash}X@{}}
\toprule
Resource & Execution / interaction examples & Copy and restore access \\
\midrule
Prompt & \texttt{\_\_call\_\_}, \texttt{get\_system\_message}, \texttt{get\_agent\_message} & Declared by server and context manager \\
Agent & \texttt{\_\_call\_\_}, \texttt{function\_callings}, \texttt{get\_schema} & Declared by server and context manager \\
Tool & \texttt{\_\_call\_\_}, \texttt{function\_callings}, \texttt{get\_schema} & Declared by server and context manager \\
Environment & \texttt{\_\_call\_\_}, \texttt{get\_state}, \texttt{get\_schema} & Declared by server and context manager \\
Memory & \texttt{start\_session}, \texttt{add\_event}, \texttt{get\_state} & Declared by context manager, not directly by server \\
\bottomrule
\end{tabularx}
\end{table}

The conceptual names \texttt{retrieve}, \texttt{get\_variables}, \texttt{set\_variables}, and the four serialization operations in Table~\ref{appx_tab:context-manager-interface} are not natively declared by these five server/context pairs; an integration must supply the corresponding adapters before advertising those capabilities. In particular, Memory restoration goes through its context manager rather than a server method.

\subsubsection{Infrastructure Services}

RSPL further includes cross-cutting services that support resource access, serialization, traceability, and retained-version recovery:

\textbf{Model manager.} Routes calls by an explicit model name or configured role, with retry handling, a configured fallback, and usage logging.

\textbf{Version manager.} Tracks resource version identifiers and lineage metadata. Recovery additionally requires retained configurations and implementation artifacts in the relevant resource manager or artifact store. A version string alone does not ensure an immutable snapshot or a failure-atomic update; these properties depend on the storage and activation implementation.

\textbf{Dynamic manager.} Loads implementation classes from source strings or files into Python modules and supports schema/argument conversion. Resource managers use these facilities when constructing objects from records. Loading a new class does not by itself migrate existing instance state or rebind every consumer; activation and persistence remain separate management responsibilities.

\textbf{Trace manager.} Records execution observations and links them to update decisions. Raw traces are not used for training or dataset synthesis by default; deployments must decide which fields to capture, redact, retain, and delete.

\paragraph{Recovery consistency requirements.}
A complete restoration must realign the active configuration, the version metadata, and subsequent consumer resolution; restoring only one does not establish a consistent recovered state. Similarly, an interrupted update must not expose a candidate merely because its configuration was assigned before version registration completed. Version identities must preserve their associated retained content if they are to serve as restoration targets.

\subsection{Layer 2: Self-Evolution Protocol Layer}
\label{sec:appendix-self-evolution-protocol-layer}

SEPL composes proposal, evaluation, and commit operations over the resource interface. The reflection instantiation below is used in the main experiments. The subsequent textual-feedback and reward-informed adapters illustrate alternative mappings; they remain unevaluated extensions rather than evidence of optimizer-independent performance or standard RL training.

\subsubsection{Evolvable Variables}

\emph{Variable lifting} resolves each resource entity to editable fields of its registration record, such as an implementation descriptor $\eta$, instantiation parameters $\theta$, or a prompt template. The entity in $\mathcal{V}_{\text{evo}}$ identifies the target; it does not limit editing to the five fields of the entity tuple. The evolvability marker selects eligible resources, and the manager specifies the permitted field projection. Identity, version bookkeeping, and edit permissions are not unrestricted candidate fields. This representation supports heterogeneous discrete edits without assuming a differentiable parameter space.

\begin{definition}[Evolvable Variable Set]
We define the universal set of evolvable variables as
\begin{equation}
\mathcal{V}_{\text{evo}} = \Bigl(\bigcup_{\tau \in \Tau} \mathcal{E}_{\tau}\Bigr) \cup \{y\},
\end{equation}
where $\mathcal{E}_{\tau}$ denotes managed resources and $y$ denotes the current task-local output/solution. Traces remain observations, not mutable evidence. A marker $g_v\in\{0,1\}$ identifies the editable subset
\begin{equation}
\Theta = \{v \in \mathcal{V}_{\text{evo}} \mid g_{v}=1\}.
\end{equation}
\end{definition}

Let $\mathfrak{V}$ denote the space of admissible resource/output states, including states with newly registered instances. An iteration state $\mathcal{V}_{\text{evo}}^{(t)}\in\mathfrak{V}$ records the current values; it is not the operator's domain by itself. Evolvability markers restrict edit targets through the manager, not operating-system permissions. Frozen implementations may still execute, and ordinary episode-state changes are distinct from implementation updates.

For example, adding an event to Memory changes stored episode data, whereas revising its retention policy changes a managed resource configuration. Likewise, advancing an Environment changes its current state, whereas revising transition rules changes its implementation. The former operations do not automatically create SEPL versions. Restoring a policy or rule implementation also does not rewind episode data or external effects unless those states were separately captured and restored. Creating a new resource requires a permitted registration operation; an editable target flag is not blanket authorization to execute generated code, access files, or transmit data.

\subsubsection{Operator Composition}

\begin{definition}[SEPL Operator]
\label{appx_def:sepl-operator}
Let $\mathfrak{V}$ be the state space and $\mathcal{P}$ the message space carrying execution evidence, proposed edits, provisional candidates, and evaluation outcomes. A \emph{SEPL operator} has the interface
\begin{equation}
f: \mathfrak{V} \times \mathcal{P}_{\text{in}} \;\rightarrow\; \mathfrak{V} \times \mathcal{P}_{\text{out}},
\end{equation}
where $\mathcal{P}_{\text{in}},\mathcal{P}_{\text{out}}\subseteq\mathcal{P}$ denote message types. An invocation returns a state $\mathcal{V}'_{\text{evo}}\in\mathfrak{V}$ and a message $p'\in\mathcal{P}_{\text{out}}$, not the spaces themselves. Composition requires compatible types and preconditions. The notation describes dataflow, not deterministic proposals; traceability and restoration additionally require recorded transitions and retained artifacts.
\end{definition}

Operators use five auxiliary spaces: traces $\mathcal{Z}$, diagnostic hypotheses $\mathcal{H}$, proposed modifications $\mathcal{D}$, objective specifications $\mathcal{G}$, and evaluation outcomes $\mathcal{S}$. The reflection instantiation composes $\{\rho,\sigma,\iota,\varepsilon,\kappa\}$ for \textsc{Reflect}, \textsc{Select}, \textsc{Improve}, \textsc{Evaluate}, and \textsc{Commit}. Hypotheses are proposed explanations of observed failures; evaluation outcomes record the results of the declared checks.

\paragraph{Retained state and candidate payload.}
For the reflection instantiation, write $V$ for the retained resource/output state and $\widetilde V$ for a provisional candidate. A message may carry the candidate, its base resource/version references, and its assessment without replacing $V$. The proposal and evaluation wrappers therefore pass $V$ through unchanged. Using $p_D$ for a message containing selected edits and $p_{\widetilde V,S}$ for a candidate-assessment message, representative wrappers are
\begin{equation}
\begin{aligned}
f_{\iota}(V,p_D) &= (V,p_{D,\widetilde V}),\qquad \widetilde V=\iota(V,D),\\
f_{\kappa}(V,p_{\widetilde V,S}) &=
\begin{cases}
(\kappa(\widetilde V,S),p_S), & \mathrm{Accept}(S),\\
(V,p_S), & \text{otherwise}.
\end{cases}
\end{aligned}
\end{equation}
Here message subscripts identify relevant contents, not an exhaustive wire format; $p_S$ retains the outcome for subsequent assessment history. Primitive signatures below abbreviate candidate computations, whereas the wrappers specify how those computations compose with the retained state. Sandboxing evaluation code is left to the deployment.

\subsubsection{Evolutionary Loop}

Initial execution under $\mathcal{V}_{\text{evo}}^{(0)}\in\mathfrak{V}$ produces a trace $Z^{(0)}\in\mathcal{Z}$, carried by $p^{(0)}\in\mathcal{P}$. Each iteration composes compatible operators:
\begin{equation}
\bigl(\mathcal{V}_{\text{evo}}^{(t+1)},\, p^{(t+1)}\bigr) = (f_n \circ \cdots \circ f_1)\bigl(\mathcal{V}_{\text{evo}}^{(t)},\, p^{(t)}\bigr),
\end{equation}
where the reflection cycle includes proposal, evaluation, commit, and task-dependent continuation under the retained state. Algorithm~\ref{alg:sepl-loop-appx} abbreviates the state as $V$ and counts update iterations after initial execution, not total model calls. It returns the retained state for early stopping or zero iterations. In contrast, the opportunity-controlled experiment permits at most three primary attempts; the generic management loop is not its exact call schedule.

Reflection produces hypotheses from traces, selection chooses permitted targets, and improvement constructs a candidate. Evaluation applies the task-specific feedback policy; commit records an accepted resource version or updates the task-local solution. Rejection must preserve the active state. Restoration targets retained resources; solution editing only revises task-local $y$.

\subsubsection{Reflection Optimizer}

\textbf{Evolvable Variables.}
The reflection optimizer edits RSPL-managed resources and, when enabled, the current task-local solution $y$. The evolvability mask restricts the selected targets while frozen implementations remain available for execution. Execution traces supply diagnostic evidence and are not retroactively modified as optimization variables.

\textbf{Operator interfaces.}
We instantiate SEPL with the canonical reflection-driven operator suite. The signatures below abbreviate primitive computations; their outputs are carried by the wrappers above rather than unconditionally replacing the retained state. Their intended roles are as follows.

\begin{itemize}[leftmargin=1em]
  \item \textbf{Reflect ($\rho$).} $\rho:\mathcal{Z}\times\mathfrak{V}\times\mathcal{S}^{*}\rightarrow\wp(\mathcal{H})$ maps execution observations, the retained state, and prior candidate assessments to hypotheses. Here $\mathcal{S}^{*}$ denotes finite sequences of assessment records, including rejected candidates. These hypotheses are proposed explanations, not established causal attributions or numerical gradients.
  \item \textbf{Select ($\sigma$).} $\sigma: \mathfrak{V} \times \wp(\mathcal{H}) \rightarrow \wp(\mathcal{D})$ selects targets implicated by the hypotheses and proposes modifications $D\subseteq\mathcal{D}$, restricted to the editable subset $\Theta$. Target selection does not itself activate a resource change.
  \item \textbf{Improve ($\iota$).} $\iota: \mathfrak{V} \times \wp(\mathcal{D}) \rightarrow \mathfrak{V}$ applies proposed edits $D$ to construct a provisional candidate. The candidate remains distinct from the active state until the declared acceptance condition is checked.
  \item \textbf{Evaluate ($\varepsilon$).} $\varepsilon:\mathfrak{V}\times\mathcal{G}\rightarrow\mathcal{S}$ evaluates a candidate under a declared task-specific feedback policy. In the label-free reasoning setting, $\mathcal{S}$ contains self-assessed satisfaction and a rationale, not benchmark correctness or a universal safety certificate.
  \item \textbf{Commit ($\kappa$).} $\kappa:\mathfrak{V}\times\mathcal{S}\rightarrow\mathfrak{V}$ records an accepted resource version or updates task-local $y$. It is invoked only after acceptance; rejection must preserve the active state. A satisfaction gate can be wrong and does not enforce true-performance monotonicity. Restoring a previously active version is a separate operation over retained state.
\end{itemize}

\textbf{The Evolutionary Loop.}
Algorithm~\ref{alg:sepl-loop-appx} composes the five operators. Initial execution produces a trace $Z\in\mathcal{Z}$ of tool outputs, decisions, failures, and progress. Reflection also receives the assessment history $F$, so a rejected candidate can inform a later proposal without replacing the active resource or rewriting the execution trace. Improvement constructs a candidate and evaluation supplies the declared checks; commit retains an accepted edit. Stopping is checked before any further task execution. The retained version history identifies potential restoration targets but does not itself demonstrate successful recovery.

\begin{appendixalgorithm}{Reflection-driven resource update (management specification)}
\label{alg:sepl-loop-appx}
\begin{algorithmic}[1]
\REQUIRE System $A$, objective/feedback policy $\mathcal{G}$, maximum update iterations $T$
\ENSURE Retained resource/output state $V$
\STATE $V \leftarrow \mathrm{ExposeResourceState}(A)$ \COMMENT{include $y$ and edit markers}
\STATE $(V,Z) \leftarrow \mathrm{Execute}(A,V)$ \COMMENT{refresh task-local $y$ and trace}
\STATE $F\leftarrow[\,]$ \COMMENT{candidate assessments, separate from trace $Z$}
\FOR{$t=1,\ldots,T$}
  \IF{$\mathrm{Stop}(V,Z,F)$}
    \STATE \textbf{break}
  \ENDIF
  \STATE $H\leftarrow \mathrm{Reflect}(Z,V,F)$
  \STATE $D\leftarrow \mathrm{Select}(V,H)$ \COMMENT{respect evolvability markers}
  \STATE $\widetilde V\leftarrow \mathrm{ImproveCandidate}(V,D)$
  \STATE $S\leftarrow \mathrm{EvaluateCandidate}(\widetilde V,\mathcal{G})$
  \IF{$\mathrm{Accept}(S)$}
    \STATE $V\leftarrow \mathrm{Commit}(\widetilde V,S)$
  \ELSE
    \STATE Keep $V$ unchanged; record candidate rejection
  \ENDIF
  \STATE $F\leftarrow\mathrm{AppendAssessment}(F,S)$
  \IF{$\mathrm{Stop}(V,Z,F)$ or $t=T$}
    \STATE \textbf{break}
  \ENDIF
  \IF{$\mathrm{NeedsExecution}(V,D,S)$}
    \STATE $(V,Z)\leftarrow \mathrm{ContinueTask}(A,V,Z)$
  \ENDIF
\ENDFOR
\STATE \textbf{return} $V$
\end{algorithmic}
\end{appendixalgorithm}

\paragraph{Continuation and stopping semantics.}
The objective/feedback policy defines \textsc{Stop} and \textsc{NeedsExecution}. Stopping concerns the retained state, not an uncommitted candidate. A completed solution-only edit retains its accepted $y$; it does not trigger another unconstrained answer generation. When resource changes require further execution, \textsc{ContinueTask} resumes the task using the retained resources and treats any accepted solution edit as part of its input. Only actual execution appends execution observations; candidate assessments remain separate diagnostic records.

The routine returns a managed state rather than guaranteeing task completion. A caller must schedule any remaining task execution within its overall budget. Candidate evaluation, proposal generation, stopping checks, and continuation can incur model or tool calls and must be included in accounting. The candidate/active-state boundary does not undo completed external actions.

\FloatBarrier
\subsubsection{Textual-feedback Adapter}
\label{app:text-feedback-adapter}
\paragraph{Scope.} This adapter is inspired by TextGrad~\citep{yuksekgonul2025optimizing}; distributing a shared critique to prompt variables is not evidence of reproducing every part of the original method. It is not used to establish the reported reflection results. The sketch synchronizes edits without a post-edit acceptance gate and therefore does not implement the complete evaluation-gated lifecycle.

\textbf{Evolvable Variables.}
In this proposed adapter, optimizable modules are represented by prompt-valued TextGrad variables with role descriptions. Each value contains the current prompt text; its role describes the module's function and guides the textual rewrite. Restricting this sketch to prompts is an adapter choice, not a limitation of TextGrad's support for other variables in a computation graph.

\textbf{Operator interfaces.}
The signatures in these extension sketches abbreviate invocations by listing their arguments; resource/output states lie in $\mathfrak{V}$, and sampled traces lie in $\mathcal{Z}$. Our mapping uses five operators: \emph{Execute}, \emph{Loss}, \emph{Backward}, \emph{Improve}, and \emph{Commit}. Here the gradient is a natural-language critique, not a numeric derivative. Commit is our synchronization step for RSPL-managed prompts, not an interface specified by TextGrad. Sharing a critique provides neither graph-specific credit assignment nor an acceptance gate.
\begin{itemize}[leftmargin=1em]
  \item \textbf{Execute ($\chi_{\mathrm{tg}}$).} $\chi_{\mathrm{tg}}: (A, \mathcal{V}_{\text{evo}}, x, f) \rightarrow \mathcal{Z}$ runs the agent under the current prompt variables and produces an execution trace/outcome.
  \item \textbf{Loss ($\lambda_{\mathrm{tg}}$).} $\lambda_{\mathrm{tg}}: \mathcal{Z} \rightarrow \mathcal{G}_{\mathrm{tg}}$, where $\mathcal{G}_{\mathrm{tg}}$ is a space of natural-language critiques (textual gradients). In this adapter, $\lambda_{\mathrm{tg}}$ is realized by \texttt{TextLoss}, which queries an evaluator LLM for feedback.
  \item \textbf{Backward ($\beta_{\mathrm{tg}}$).} $\beta_{\mathrm{tg}}: \mathfrak{V} \times \mathcal{G}_{\mathrm{tg}} \rightarrow \mathfrak{V}$ assigns textual gradients to optimizable prompt variables by storing the critique (optionally with context) in a per-variable gradient buffer. The sketch distributes the same critique to each optimizable prompt variable.
  \item \textbf{Improve ($\iota_{\mathrm{tg}}$).} $\iota_{\mathrm{tg}}: \mathfrak{V} \rightarrow \mathfrak{V}$ rewrites prompt variables via a textual-gradient-descent step: it constructs an update instruction from each variable's role description, current value, and accumulated textual gradients, then queries an optimizer LLM and extracts the improved variable text from a constrained output format.
  \item \textbf{Commit ($\kappa_{\mathrm{tg}}$).} $\kappa_{\mathrm{tg}}: \mathfrak{V} \rightarrow \mathfrak{V}$ synchronizes the updated prompt variables back into the running agent and clears caches, synchronizing the edited fields with the managed resource configuration.
\end{itemize}

\textbf{The Evolutionary Loop.}
Algorithm~\ref{alg:textgrad-loop-appx} executes the current prompts to obtain $Z\in\mathcal{Z}$ and an evaluator critique $g\in\mathcal{G}_{\mathrm{tg}}$. It checks the stopping condition on those prompts before modifying them. Otherwise, the adapter shares the critique across prompt-variable buffers, rewrites the prompts, and synchronizes the edited values before the next execution. This ordering avoids an extra rewrite after a stopping decision; it does not add a post-edit acceptance gate. The final update at budget exhaustion may therefore remain unevaluated.

\begin{appendixalgorithm}{Textual-feedback adapter sketch (without a post-edit gate)}
  \label{alg:textgrad-loop-appx}
  \newcommand{\algcommenttg}[1]{\hfill\parbox[t]{0.48\linewidth}{\raggedleft\footnotesize\textcolor{gray}{$\rhd$ \textit{#1}}}}
  \begin{algorithmic}[1]
  \renewcommand{\algorithmicrequire}{\textbf{Input:}}
  \renewcommand{\algorithmicensure}{\textbf{Output:}}
  \REQUIRE Agentic system $\mathcal{A}$, task $x$, optional attachments $f$, update budget $K\geq0$, evaluator/optimizer LLMs $M_{\text{eval}}, M_{\text{opt}}$
  \ENSURE Updated state $\mathcal{V}_{\text{evo}}^{*}$ (prompt variables updated via TextGrad)

  \STATE \textcolor{gray}{// Phase 0: Setup}
  \STATE Set backward engine to $M_{\text{eval}}$ \algcommenttg{Evaluator used by TextLoss}
  \STATE $\mathcal{V}_{\text{evo}}^{(0)} \leftarrow \text{VariableLifting}(\mathcal{A})$ \algcommenttg{Lift optimizable prompts to TextGrad variables}
  \STATE $\ell \leftarrow 0$
  \STATE Initialize textual optimizer with $M_{\text{opt}}$ \algcommenttg{TextualGradientDescent over prompt vars}

  \vspace{0.2em}
  \STATE \textcolor{gray}{// Optimization Cycle}
  \FOR{$k = 0, 1, \ldots, K-1$}
      \STATE \textcolor{gray}{// Phase 1: Execute (Forward)}
      \STATE $Z^{(k)} \leftarrow \chi_{\mathrm{tg}}(\mathcal{A}, \mathcal{V}_{\text{evo}}^{(k)}, x, f)$ \algcommenttg{Run agent with current prompts}
      
      \vspace{0.1em}
      \STATE \textcolor{gray}{// Phase 2: Loss (Textual Gradient)}
      \STATE Build evaluation instruction from $Z^{(k)}$ \algcommenttg{Condition on success/error}
      \STATE $g^{(k)} \leftarrow \lambda_{\mathrm{tg}}(Z^{(k)})$ \algcommenttg{TextLoss produces critique string}
      \IF{$\text{Converged}(g^{(k)})$}
          \STATE \textbf{break}
      \ENDIF
      
      \vspace{0.1em}
      \STATE \textcolor{gray}{// Phase 3: Backward (Assign Gradients)}
      \STATE $\mathcal{V}_{\text{evo}}^{(k)} \leftarrow \beta_{\mathrm{tg}}(\mathcal{V}_{\text{evo}}^{(k)}, g^{(k)})$ \algcommenttg{Assign critique to gradient buffers}
      
      \vspace{0.1em}
      \STATE \textcolor{gray}{// Phase 4: Improve (Textual Gradient Descent)}
      \STATE $\widetilde{\mathcal{V}}_{\text{evo}}^{(k+1)} \leftarrow \iota_{\mathrm{tg}}(\mathcal{V}_{\text{evo}}^{(k)})$ \algcommenttg{Rewrite prompts via textual GD}
      
      \vspace{0.1em}
      \STATE \textcolor{gray}{// Phase 5: Commit \& Next Iteration}
      \STATE $\mathcal{V}_{\text{evo}}^{(k+1)} \leftarrow \kappa_{\mathrm{tg}}(\widetilde{\mathcal{V}}_{\text{evo}}^{(k+1)})$ \algcommenttg{Sync back; clear caches}
      \STATE $\ell \leftarrow k+1$
  \ENDFOR
  \STATE \textbf{return} $\mathcal{V}_{\text{evo}}^{(\ell)}$
  \end{algorithmic}
  \end{appendixalgorithm}

\FloatBarrier
\subsubsection{Reward-informed Editing Sketch (Reinforce++-inspired)}
\label{app:reward-editing}
\paragraph{Scope and supervision.} This is a heuristic resource-editing sketch, not standard Reinforce++ training~\citep{hu2025reinforce++}. Text similarity is not a token-probability ratio, and no policy-gradient update is defined. The labeled inputs $y^*$ and $y_{\mathrm{sft}}$ belong to a separate supervised setting, not the label-free GPQA-Diamond/AIME reflection results. The sketch also lacks a separate post-edit acceptance evaluation. It is retained as an unvalidated extension description, not empirical evidence.

\textbf{Evolvable Variables.}
The sketch uses reward-derived diagnostics to propose edits to selected prompt and tool resources, with optional revision of the current solution. These diagnostics condition the editor rather than update model parameters through policy gradients.

\textbf{Operator interfaces.}
The following heuristic combines a clipped score and a reference-text penalty to condition an editor. These quantities should not be interpreted as a likelihood-ratio objective for a trainable language-model policy.

Let $\eta(a,b)\in[0,1]$ be a declared text-similarity measure, with finite task rewards, penalty weight $\beta\geq0$, clipping parameter $0\leq\epsilon<1$, and numerical floor $0<\epsilon_0\leq1$. Update budgets are nonnegative integers. These specify the admissible inputs, not fitted experimental hyperparameters. The same notation applies to the group-based sketch below.

The editable-resource projection $\mathfrak{V}_{\mathrm{edit}}$ contains only enabled persistent fields; \textsc{GetEditable} excludes task-local $y$. The editor's projection space $\mathfrak{U}$ contains either such a resource projection or a task-local solution. \textsc{ReplaceFields} and \textsc{SetOutput} insert the respective edits into the full state. This convention also governs the group-based sketch; solutions remain task-local.
\begin{itemize}[leftmargin=1em]
  \item \textbf{Sample ($\chi_{\mathrm{rpp}}$).} $\chi_{\mathrm{rpp}}: (A, \mathcal{V}_{\text{evo}}, x, f) \rightarrow \mathcal{Z}$ samples a rollout under the current resources and yields an execution trace containing the produced answer.
  \item \textbf{Reward ($\varepsilon_{\mathrm{rpp}}$).} $\varepsilon_{\mathrm{rpp}}: (y^{(t)}, y^{(t-1)}, y^{*}, y_{\mathrm{sft}}) \rightarrow (r^{(t)}, A^{(t)}, J^{(t)}, \pi^{(t)})$ computes the diagnostic signal tuple from the current solution $y^{(t)}$. Here $r^{(t)}$ is a task reward comparing $y^{(t)}$ with $y^{*}$, and $\pi^{(t)}$ is a heuristic score based on text similarity $\eta(\cdot,\cdot)$ as $\pi^{(t)} \triangleq \eta(y^{(t-1)}, y^{(t)})$ (the sketch does not use token-level probabilities). We define a penalty to a reference solution $y_{\mathrm{sft}}$ as $\mathrm{pen}^{(t)} \triangleq \beta\,\bigl|\log \max(\eta(y_{\mathrm{sft}}, y^{(t)}), \epsilon_0)\bigr|$ and set $A^{(t)} \triangleq r^{(t)} - \mathrm{pen}^{(t)}$. The heuristic clipped score is
  \[
    J^{(t)} \triangleq \min\bigl(\pi^{(t)} A^{(t)},\; \bar{\pi}^{(t)} A^{(t)}\bigr),
    \quad \bar{\pi}^{(t)} \triangleq \mathrm{clip}(\pi^{(t)}, 1-\epsilon, 1+\epsilon).
  \]
  \item \textbf{Diagnose ($\delta_{\mathrm{rpp}}$).} $\delta_{\mathrm{rpp}}: (\mathcal{Z}, \mathcal{V}_{\text{edit}}, r^{(t)}, A^{(t)}, J^{(t)}, \pi^{(t)}) \rightarrow \mathcal{H}$ produces an edit-oriented diagnosis that is explicitly conditioned on the reward-derived metrics and the execution trace.
  \item \textbf{Improve ($\iota_{\mathrm{rpp}}$).} $\iota_{\mathrm{rpp}}: \mathfrak{U}\times\mathcal{H}\rightarrow\mathfrak{U}$ edits the selected resource projection or, when enabled, the task-local solution. The output retains the input projection's type; unselected fields remain unchanged.
  \item \textbf{Commit ($\kappa_{\mathrm{rpp}}$).} $\kappa_{\mathrm{rpp}}: \mathfrak{V}_{\mathrm{edit}} \rightarrow \mathfrak{V}_{\mathrm{edit}}$ synchronizes proposed persistent edits with RSPL without a separate post-edit acceptance test. Task-local solution edits use \textsc{SetOutput} instead.
\end{itemize}

\textbf{The Evolutionary Loop.}
Algorithm~\ref{alg:reinforcepp-loop-appx} organizes the heuristic into reward computation, resource editing, re-execution, optional solution revision, and early stopping. The clipped similarity score and reference-text penalty provide diagnostic inputs to the editor. No likelihood-ratio estimate or policy-gradient update is defined. Optional solution edits update $y$ in the task state, not a persistent registry entry; the recorded execution trace remains unchanged.

The reward tuple at iteration $t$ evaluates the pre-edit solution $y^{(t)}$. When reused after a resource update, it is historical diagnostic context, not a score for the newly generated or rewritten solution. The declared \textsc{Satisfied} check must assess the current $y^{(t+1)}$; earlier trace success alone cannot certify a later text edit. Any new model or execution calls made by this check count toward the overall cost, even though $T$ counts update iterations only.

\begin{appendixalgorithm}{Heuristic reward-informed editing sketch (not standard Reinforce++)}
  \label{alg:reinforcepp-loop-appx}
  \newcommand{\algcommentrpp}[1]{\hfill\parbox[t]{0.48\linewidth}{\raggedleft\footnotesize\textcolor{gray}{$\rhd$ \textit{#1}}}}
  \begin{algorithmic}[1]
  \renewcommand{\algorithmicrequire}{\textbf{Input:}}
  \renewcommand{\algorithmicensure}{\textbf{Output:}}
  \REQUIRE Agentic system $\mathcal{A}$, task $x$, optional attachments $f$, ground truth $y^{*}$, reference solution $y_{\mathrm{sft}}$, update budget $T\geq0$
  \ENSURE Final solution $y^{(\ell)}$ and updated editable resources $\mathcal{V}_{\text{edit}}$

  \STATE \textcolor{gray}{// Initialization}
  \STATE $\mathcal{V}_{\text{evo}}^{(0)} \leftarrow \text{VariableLifting}(\mathcal{A})$ \algcommentrpp{Expose editable resources}
  \STATE $Z^{(0)} \leftarrow \chi_{\mathrm{rpp}}(\mathcal{A}, \mathcal{V}_{\text{evo}}^{(0)}, x, f)$ \algcommentrpp{Sample once}
  \STATE Extract solution $y^{(0)}$ from $Z^{(0)}$
  \STATE $\mathcal{V}_{\text{evo}}^{(0)} \leftarrow \operatorname{SetOutput}(\mathcal{V}_{\text{evo}}^{(0)},y^{(0)})$
  \STATE $\ell \leftarrow 0$
  \STATE $y^{(-1)} \leftarrow y^{(0)}$ \algcommentrpp{Initialize previous solution}
  
  \FOR{$t = 0, 1, \ldots, T-1$}
      \STATE \textcolor{gray}{// Phase 1: Reward and heuristic similarity score}
      \STATE $(r^{(t)}, A^{(t)}, J^{(t)}, \pi^{(t)}) \leftarrow \varepsilon_{\mathrm{rpp}}(y^{(t)}, y^{(t-1)}, y^{*}, y_{\mathrm{sft}})$

      \vspace{0.1em}
      \STATE \textcolor{gray}{// Phase 2: Improve editable resources (prompt and tool)}
      \STATE $\mathcal{V}_{\text{edit}}^{(t)} \leftarrow \text{GetEditable}(\mathcal{V}_{\text{evo}}^{(t)})$
      \STATE $H_{\text{edit}}^{(t)} \leftarrow \delta_{\mathrm{rpp}}(Z^{(t)}, \mathcal{V}_{\text{edit}}^{(t)}, r^{(t)}, A^{(t)}, J^{(t)}, \pi^{(t)})$
      \STATE $\widetilde{\mathcal{V}}_{\text{edit}}^{(t+1)} \leftarrow \iota_{\mathrm{rpp}}(\mathcal{V}_{\text{edit}}^{(t)}, H_{\text{edit}}^{(t)})$
      \STATE $\mathcal{V}_{\text{edit}}^{(t+1)} \leftarrow \kappa_{\mathrm{rpp}}(\widetilde{\mathcal{V}}_{\text{edit}}^{(t+1)})$ \algcommentrpp{Commit updates}
      \STATE $\mathcal{V}_{\text{evo}}^{(t+1)} \leftarrow \operatorname{ReplaceFields}(\mathcal{V}_{\text{evo}}^{(t)},\mathcal{V}_{\text{edit}}^{(t+1)})$

      \vspace{0.1em}
      \STATE \textcolor{gray}{// Phase 3: Re-execute under updated resources}
      \STATE $Z^{(t+1)} \leftarrow \chi_{\mathrm{rpp}}(\mathcal{A}, \mathcal{V}_{\text{evo}}^{(t+1)}, x, f)$
      \STATE Extract solution $y^{(t+1)}$ from $Z^{(t+1)}$

      \vspace{0.1em}
      \STATE \textcolor{gray}{// Phase 4: Optional solution refinement}
      \IF{solution refinement is enabled}
      \STATE $H_{\text{sol}}^{(t)} \leftarrow \delta_{\mathrm{rpp}}(Z^{(t+1)}, \{y^{(t+1)}\}, r^{(t)}, A^{(t)}, J^{(t)}, \pi^{(t)})$
      \STATE $\widetilde{y}^{(t+1)} \leftarrow \iota_{\mathrm{rpp}}(y^{(t+1)}, H_{\text{sol}}^{(t)})$ \algcommentrpp{Edit solution text (candidate)}
      \STATE $y^{(t+1)} \leftarrow \widetilde{y}^{(t+1)}$ \algcommentrpp{Accept task-local edit}
      \ENDIF
      \STATE $\mathcal{V}_{\text{evo}}^{(t+1)} \leftarrow \operatorname{SetOutput}(\mathcal{V}_{\text{evo}}^{(t+1)},y^{(t+1)})$
      
      \vspace{0.1em}
      \STATE \textcolor{gray}{// Phase 5: Early stopping}
      \STATE $\ell \leftarrow t+1$
      \IF{$\text{Satisfied}(Z^{(t+1)}, y^{(t+1)})$}
          \STATE \textbf{break}
      \ENDIF
  \ENDFOR
  \STATE \textbf{return} $(y^{(\ell)},\,\operatorname{GetEditable}(\mathcal{V}_{\text{evo}}^{(\ell)}))$
  \end{algorithmic}
  \end{appendixalgorithm}

\FloatBarrier
\subsubsection{Group-normalized Reward Editing Sketch (GRPO-inspired)}
\label{app:group-editing}
\paragraph{Scope and supervision.} This is a heuristic using group-normalized reward summaries to condition an editor, not standard GRPO optimization~\citep{shao2024deepseekmath}. It shares the limitations of Appendix~\ref{app:reward-editing}: labeled feedback, similarity rather than likelihood ratios, no defined policy-gradient update, and no separate post-edit gate. It does not produce the main reflection results.

\textbf{Evolvable Variables.}
The group-based sketch proposes edits to selected prompt and tool resources and, optionally, the current solution. It compares labeled rewards across sampled candidates to provide diagnostic context for the editor, without differentiating through a language-model policy.

\textbf{Operator interfaces.}
The operators below sample candidate solutions, normalize their labeled rewards, and pass the summaries to an editor. Their signatures describe how the heuristic interacts with the resource interface; they do not define a GRPO policy-gradient objective.

For the update loop, $K\geq2$ is an integer; the sampling interface also permits one rollout for initialization and re-execution. We use the population moments $\bar r=K^{-1}\sum_i r_i$ and $\sigma_r^2=K^{-1}\sum_i(r_i-\bar r)^2$. If every reward is equal, the advantages are zero rather than undefined. Appendix~\ref{app:reward-editing} defines the shared similarity parameters and projection types, not experimental settings.
\begin{itemize}[leftmargin=1em]
  \item \textbf{Sample ($\chi_{\mathrm{grpo}}$).} $\chi_{\mathrm{grpo}}: (A, \mathcal{V}_{\text{evo}}, x, f, K) \mapsto \{Z_i\}_{i=1}^K$ samples $K$ independent rollouts under the current resources, yielding $K$ execution traces each containing a candidate solution $y_i$.
  \item \textbf{Reward ($\varepsilon_{\mathrm{grpo}}$).} $\varepsilon_{\mathrm{grpo}}: (\{y_i\}_{i=1}^K, y^{*}, y^{(t-1)}) \rightarrow (\{r_i\}_{i=1}^K, \{A_i\}_{i=1}^K, \{J_i\}_{i=1}^K, \{\pi_i\}_{i=1}^K)$ computes reward-derived diagnostic signals for all $K$ candidates. For each candidate $y_i$, we compute a task reward $r_i$ comparing $y_i$ with $y^{*}$, a heuristic similarity score $\pi_i \triangleq \eta(y^{(t-1)}, y_i)$ approximated via text similarity $\eta(\cdot,\cdot)$ (the sketch does not use token-level probabilities), and a group normalized advantage $A_i$ by normalizing rewards across the candidate set: $A_i = (r_i - \bar{r}) / \max(\sigma_r,\epsilon_0)$ where $\bar{r}$ and $\sigma_r$ are the mean and standard deviation of $\{r_i\}_{i=1}^K$, and $\epsilon_0>0$ avoids division by zero. The heuristic clipped score for each candidate is
  \[
    J_i \triangleq \min\bigl(\pi_i A_i,\; \bar{\pi}_i A_i\bigr),
    \quad \bar{\pi}_i \triangleq \begin{cases}
      \min(\pi_i, 1+\epsilon) & \text{if } A_i \geq 0 \\
      \max(\pi_i, 1-\epsilon) & \text{if } A_i < 0
    \end{cases}.
  \]
  \item \textbf{Diagnose ($\delta_{\mathrm{grpo}}$).} $\delta_{\mathrm{grpo}}: (\{Z_i\}_{i=1}^K, \mathcal{V}_{\text{edit}}, \{r_i, A_i, J_i, \pi_i\}_{i=1}^K) \rightarrow \mathcal{H}$ produces an edit-oriented diagnosis that is explicitly conditioned on the multiple candidate solutions and their reward-derived metrics, enabling the optimizer to identify patterns across candidates.
  \item \textbf{Improve ($\iota_{\mathrm{grpo}}$).} $\iota_{\mathrm{grpo}}: \mathfrak{U}\times\mathcal{H}\rightarrow\mathfrak{U}$ edits the selected resource projection or, when enabled, the task-local solution, preserving the projection type and all unselected fields.
  \item \textbf{Commit ($\kappa_{\mathrm{grpo}}$).} $\kappa_{\mathrm{grpo}}: \mathfrak{V}_{\mathrm{edit}} \rightarrow \mathfrak{V}_{\mathrm{edit}}$ synchronizes proposed persistent edits with RSPL without a separate post-edit acceptance test. Task-local solution edits use \textsc{SetOutput} instead.
\end{itemize}

\textbf{The Evolutionary Loop.}
Algorithm~\ref{alg:grpo-loop-appx} samples $K$ candidate solutions, constructs group-normalized reward summaries, and conditions resource edits on their traces. It then re-executes the updated resources, optionally revises the solution, and checks the stopping criterion. The group statistics guide the heuristic editor rather than a gradient update. They remain attached to the sampled pre-edit candidates, not to the later solution. As in Appendix~\ref{app:reward-editing}, solution edits update task-local $y$ without altering prior observations, and stopping must assess the current solution.

\begin{appendixalgorithm}{Heuristic group-reward editing sketch (not standard GRPO)}
  \label{alg:grpo-loop-appx}
  \newcommand{\algcommentgrpo}[1]{\hfill\parbox[t]{0.48\linewidth}{\raggedleft\footnotesize\textcolor{gray}{$\rhd$ \textit{#1}}}}
  \begin{algorithmic}[1]
  \renewcommand{\algorithmicrequire}{\textbf{Input:}}
  \renewcommand{\algorithmicensure}{\textbf{Output:}}
  \REQUIRE Agentic system $\mathcal{A}$, task $x$, optional attachments $f$, ground truth $y^{*}$, update budget $T\geq0$, candidates per group $K\geq2$
  \ENSURE Final solution $y^{(\ell)}$ and updated editable resources $\mathcal{V}_{\text{edit}}$

  \STATE \textcolor{gray}{// Initialization}
  \STATE $\mathcal{V}_{\text{evo}}^{(0)} \leftarrow \text{VariableLifting}(\mathcal{A})$ \algcommentgrpo{Expose editable resources}
  \STATE $\{Z^{(0)}\} \leftarrow \chi_{\mathrm{grpo}}(\mathcal{A}, \mathcal{V}_{\text{evo}}^{(0)}, x, f, 1)$ \algcommentgrpo{Sample initial solution}
  \STATE Extract solution $y^{(0)}$ from $Z^{(0)}$
  \STATE $\mathcal{V}_{\text{evo}}^{(0)} \leftarrow \operatorname{SetOutput}(\mathcal{V}_{\text{evo}}^{(0)},y^{(0)})$
  \STATE $\ell \leftarrow 0$
  \STATE $y^{(-1)} \leftarrow y^{(0)}$ \algcommentgrpo{Initialize previous solution}
  
  \FOR{$t = 0, 1, \ldots, T-1$}
      \STATE \textcolor{gray}{// Phase 1: Sample multiple candidates}
      \STATE $\{Z_i^{(t)}\}_{i=1}^K \leftarrow \chi_{\mathrm{grpo}}(\mathcal{A}, \mathcal{V}_{\text{evo}}^{(t)}, x, f, K)$ \algcommentgrpo{Sample $K$ rollouts}
      \STATE Extract candidate solutions $\{y_i^{(t)}\}_{i=1}^K$ from $\{Z_i^{(t)}\}_{i=1}^K$
      
      \vspace{0.1em}
      \STATE \textcolor{gray}{// Phase 2: Group rewards and heuristic scores}
      \STATE $(\{r_i^{(t)}\}_{i=1}^K, \{A_i^{(t)}\}_{i=1}^K, \{J_i^{(t)}\}_{i=1}^K, \{\pi_i^{(t)}\}_{i=1}^K) \leftarrow \varepsilon_{\mathrm{grpo}}(\{y_i^{(t)}\}_{i=1}^K, y^{*}, y^{(t-1)})$

      \vspace{0.1em}
      \STATE \textcolor{gray}{// Phase 3: Improve editable resources (prompt and tool)}
      \STATE $\mathcal{V}_{\text{edit}}^{(t)} \leftarrow \text{GetEditable}(\mathcal{V}_{\text{evo}}^{(t)})$
      \STATE $H_{\text{edit}}^{(t)} \leftarrow \delta_{\mathrm{grpo}}(\{Z_i^{(t)}\}_{i=1}^K, \mathcal{V}_{\text{edit}}^{(t)}, \{r_i^{(t)}, A_i^{(t)}, J_i^{(t)}, \pi_i^{(t)}\}_{i=1}^K)$
      \STATE $\widetilde{\mathcal{V}}_{\text{edit}}^{(t+1)} \leftarrow \iota_{\mathrm{grpo}}(\mathcal{V}_{\text{edit}}^{(t)}, H_{\text{edit}}^{(t)})$
      \STATE $\mathcal{V}_{\text{edit}}^{(t+1)} \leftarrow \kappa_{\mathrm{grpo}}(\widetilde{\mathcal{V}}_{\text{edit}}^{(t+1)})$ \algcommentgrpo{Commit updates}
      \STATE $\mathcal{V}_{\text{evo}}^{(t+1)} \leftarrow \operatorname{ReplaceFields}(\mathcal{V}_{\text{evo}}^{(t)},\mathcal{V}_{\text{edit}}^{(t+1)})$

      \vspace{0.1em}
      \STATE \textcolor{gray}{// Phase 4: Re-execute under updated resources}
      \STATE $\{Z^{(t+1)}\} \leftarrow \chi_{\mathrm{grpo}}(\mathcal{A}, \mathcal{V}_{\text{evo}}^{(t+1)}, x, f, 1)$
      \STATE Extract solution $y^{(t+1)}$ from $Z^{(t+1)}$

      \vspace{0.1em}
      \STATE \textcolor{gray}{// Phase 5: Optional solution refinement}
      \IF{solution refinement is enabled}
      \STATE $H_{\text{sol}}^{(t)} \leftarrow \delta_{\mathrm{grpo}}(\{Z_i^{(t)}\}_{i=1}^K, \{y^{(t+1)}\}, \{r_i^{(t)}, A_i^{(t)}, J_i^{(t)}, \pi_i^{(t)}\}_{i=1}^K)$
      \STATE $\widetilde{y}^{(t+1)} \leftarrow \iota_{\mathrm{grpo}}(y^{(t+1)}, H_{\text{sol}}^{(t)})$ \algcommentgrpo{Edit solution text (candidate)}
      \STATE $y^{(t+1)} \leftarrow \widetilde{y}^{(t+1)}$ \algcommentgrpo{Accept task-local edit}
      \ENDIF
      \STATE $\mathcal{V}_{\text{evo}}^{(t+1)} \leftarrow \operatorname{SetOutput}(\mathcal{V}_{\text{evo}}^{(t+1)},y^{(t+1)})$
      
      \vspace{0.1em}
      \STATE \textcolor{gray}{// Phase 6: Early stopping}
      \STATE $\ell \leftarrow t+1$
      \IF{$\text{Satisfied}(Z^{(t+1)}, y^{(t+1)})$}
          \STATE \textbf{break}
      \ENDIF
  \ENDFOR
  \STATE \textbf{return} $(y^{(\ell)},\,\operatorname{GetEditable}(\mathcal{V}_{\text{evo}}^{(\ell)}))$
  \end{algorithmic}
  \end{appendixalgorithm}

\end{document}